\pdfoutput=1
\documentclass[review]{elsarticle}
\usepackage{subfigure}
\usepackage{longtable}
\usepackage{algorithm, algorithmic}
\usepackage{bm}
\usepackage{booktabs}
\usepackage{multirow}
\usepackage{geometry}
\usepackage{amsfonts,amssymb}

\usepackage{lineno,hyperref}
\modulolinenumbers[5]

\journal{arXiv}

\begin{document}

\begin{frontmatter}

\title{Temporal Attention Augmented Transformer Hawkes Process}


\author[mymainaddress]{Lu-ning Zhang}
\ead{luning_zhang@foxmail.com}

\author[mymainaddress]{Jian-wei Liu\corref{mycorrespondingauthor}}
\cortext[mycorrespondingauthor]{Corresponding author}
\ead{liujw@cup.edu.cn}

\author[mymainaddress]{Zhi-yan Song}

\author[mymainaddress]{Xin Zuo}

\address[mymainaddress]{Department of Automation, College of Information Science and Engineering,
China University of Petroleum , Beijing, Beijing, China}

\begin{abstract}
In recent years, mining the knowledge from asynchronous sequences by Hawkes process is a subject worthy of continued attention, and Hawkes processes based on the neural network have gradually become the most hotly researched fields, especially based on the recurrence neural network (RNN). However, these models still contain some inherent shortcomings of RNN, such as vanishing and exploding gradient and long-term dependency problems. Meanwhile, Transformer based on self-attention has achieved great success in sequential modeling like text processing and speech recognition. Although the Transformer Hawkes process (THP) has gained huge performance improvement, THPs don’t effectively utilize the temporal information in the asynchronous events, for these asynchronous sequences, the event occurrence instants are as important as the types of events, while conventional THPs simply convert temporal information into position encoding and add them as the input of transformer. With this in mind, we come up with a new kind of Transformer-based Hawkes process model, Temporal Attention Augmented Transformer Hawkes Process (TAA-THP), we modify the traditional dot-product attention structure, and introduce the temporal encoding into attention structure. We conduct numerous experiments on a wide range of synthetic and real-life datasets to validate the performance of our proposed TAA-THP model, significantly improvement compared with existing baseline models on the different measurements is achieved, including log-likelihood on the test dataset, and prediction accuracies of event types and occurrence times. In addition, through the ablation studies, we vividly demonstrate the merit of introducing additional temporal attention by comparing the performance of the model with and without temporal attention.
\end{abstract}

\begin{keyword}
Hawkes process, Transformer Hawkes process, Position encoding, Temporal attention, Dot-product attention
\end{keyword}

\end{frontmatter}

\section{Introduction}

With the development of information technology, most of the engineering and science records are currently stored in computers with the electronic forms, such as earthquake records \cite{1ogata1981lewis}, medical records \cite{2johnson2016mimic}, public safety records \cite{3mohler2018improving}, failure records \cite{4zhang2020survival}, and so on. Meanwhile, with the increase of internet applications, these diverse applications also lead to a large number of records, like IPTV records \cite{5luo2014you} and social network records \cite{6zhao2015seismic}. In general, these records are in the form of asynchronous sequence data, which contain the event occurring time and type of event.

These event sequence data contain a lot of valuable knowledge, which can be used to understand the relationship between events, predict possible future events, as well as the time of occurrence, and so on. How to mine the knowledge from these asynchronous sequences is a subject worthy of continued attention, and one of the most widely used methods is the point process model \cite{7daley2007introduction}, and the most commonly used model in the point process model is the Hawkes process model \cite{8hawkes1971spectra}. 

In previous years, because the Hawkes process with self-excited characteristics can well model the trigger mode between events, the traditional Hawkes process is widely applied in many fields. For instance, Ogata utilizes space-time Hawkes process to analyze the earthquake and corresponding aftershock \cite{1ogata1981lewis}. Zhao et al. make use of the Hawkes process to predict the popularity of tweets on Internet \cite{6zhao2015seismic}. Reynaud-Bouret et al. provide a new way to detect the distances between genomic events on the DNA (Deoxyribonucleic Acid) sequences \cite{9reynaud2010adaptive}. Kobayashi and Lamhbiotte \cite{10kobayashi2016tideh} present a new framework of the time-dependent Hawkes process, whose impact functions are time-dependent. They make use of this model to reveal the periodic characteristics of Twitter reposting. Xu et al. utilize the Hawkes process to uncover the Granger causality between users choosing to watch TV programs \cite{11xu2016learning}. Zhou et al. present ADMM (Alternating Direction Method of Multipliers)-based algorithm to train the Hawkes process to find out the relationship between the social media user \cite{12zhou2013learning}.

Unfortunately, there are two critical inherent shortcomings for the traditional Hawkes process, one shortcoming is that the traditional Hawkes process ignores the influence of mutual inhibition between events, which does not conform to the actual situation to a certain extent. Another shortcoming is the lack of strong nonlinear fitting capability in the traditional Hawkes process, which also limits the expressive ability of the model. 

Therefore, to mitigate the above problem, and with the development of neural networks and deep learning, also due to the ability to effectively model sequence data of recurrent neural network (RNN), the Hawkes process models based on RNN are proposed. For example, Du et al. embed sequence data information (including time-stamp and event type) into RNN and propose recurrent marked temporal point processes (RMTPP) to model the conditional intensity function considering the historical nonlinear dependence \cite{13du2016recurrent}. Similar to RMTPP, Mei et al. propose a continuous-time LSTM (Long Short-Term Memory) to simulate the conditional intensity of point processes, which is called neural Hawkes processes \cite{14mei2017neural}. In this continuous neural Hawkes process, the effect of the previous event is decreasing with time. And in \cite{15xiao2017modeling} two RNNs are used to model the conditional intensity function, one is for processing time-stamp information and the other for processing historical event information.

The goals of these models are effectively model the sequence data, and accurately predict the types of events that will occur in the future and the moments of event occurrence. For instance, for electronic medical records, we can predict the types and times of the next attack based on the patient’s medical history, and provide more effective and timely help for the patient's treatment. We can also predict the next possible failure time and type for a large-scale production system based on the failure sequence, and perform maintenance and prevention in advance to improve safety and economic benefits for the production process.

However, these models based on RNN inevitably inherit the disadvantages of RNN. For example, for certain chronic sequelae, it may take a long time for patients to develop symptoms. These sequelae may have obvious long-term characteristics, such as diabetes, cancer, and other chronic diseases. These RNN-based models are difficult to describe the long-term dependence between events with a long sequence distance \cite{16bengio1994learning}, while the ideal point process model should be able to attack these problems. Moreover, when training RNN-based models, problems such as gradient vanishing and explosion \cite{17pascanu2013difficulty} often appear, this will necessarily impact the performance of the model.

Fortunately, RNN is not the only choice of sequential modeling now, with the advent of self-attention mechanism \cite{18DBLP:journals/corr/BahdanauCB14}, sequence models in literatures have become vast and are growing rapidly, among them, the most striking one, transformer \cite{19vaswani2017attention}, are developed and applied to speech recognition \cite{20yu2016automatic}, machine translation \cite{21koehn2009statistical}, video recognition \cite{22girdhar2019video} and other fields. Inspired by these successes, Zhang et al. present the self-attention Hawkes process \cite{23zhang2020self}, furthermore, Zuo et al. propose transformer Hawkes process based on the attention mechanism and encoder structure in transformer \cite{24DBLP:conf/icml/ZuoJLZZ20}. 

But for the existing attention Hawkes model, the input feeding into the transformer is a simple stacking of the event encoding and the temporal position encoding, while we tend to extract the existing underlying intrinsic information in the event type and the time stamp in the event sequence. Therefore, inspired by the relative encoding Hawkes process \cite{25dai2019transformer,26al2019character}, we modify the traditional dot product attention and propose a new dot product attention mechanism, which additionally introduces temporal encoding as input to the attention mechanism, we dub this model as Temporal Augmented Attention Transformer Hawkes Process (TAA-THP).

Our paper is organized as follows. In section 2, we systematically introduce the related work of our methods, and next, we describe our model in detail, which is related to the modified dot-product temporal augmented attention structure, and the acquisition of conditional intensity function based on the hidden representation of events sequences. In section 4, we instruct the prediction method and the objective function of TAA-THP. Then, we conduct the experiments to confirm the effectiveness of the TAA-THP model, we also perform the ablation study to figure out the impact of the introduction of temporal attention. At last, we summarize our research work and look forward to future research directions.

\section{Related work}

\subsection{Traditional Hawkes process}

Hawkes process is one of the most widely applied point process models in sequential modeling, which is proposed by Alan Hawkes in 1971 \cite{8hawkes1971spectra}. The general paradigm of Hawkes process is shown as Eq. \ref{eq1}:

\begin{equation}
	\label{eq1}
	\lambda (t) = \mu (t) + \sum\limits_{i:t_i  < t} {\phi (t - t_i )} 	
\end{equation}

where $\mu (t)$ is the background intensity function, describes the base possibility of event occurrence over time, $\phi (t)$ is the impact function, which is used to represents the historical event impact, and $\sum\limits_{i:t_i  < t} {\phi (t - t_i )} 	$ records all the impact of history events to this current instant $t$. The traditional Hawkes process model in Eq. \ref{eq1} ignores the inhibition effect between the events, and until now, there are still countless variants based on this model in various application fields.

Mohler proposes a novel modulated Hawkes process, to quickly identify risks to protect communities from a range of social harm events, such as crime, drug abuse, traffic accident and medical emergencies \cite{3mohler2018improving}. Zhang et al. \cite{4zhang2020survival} use the Weibull background intensity instead of the constant background intensity, which can better model the trend of failure occurrence over time, Zhang et al. verify the effectiveness of the Weibull-Hawkes process, and reveal the trend of background probability of failures in the compressor station over time and the Granger causality between the failures. Xu et al. \cite{5luo2014you} use the non-parametric Hawkes process model, and a corresponding learning algorithm to obtain the Granger causality on of IPTV users on watching the program.

Kobayashi and Lamhbiotte use the time-dependent Hawkes process to prove that Twitter reposts have a clear tidal effect \cite{10kobayashi2016tideh}. Yang et al. \cite{27yang2017online} come up with an online learning method of Hawkes process based on the nonparametric method. In 2018, Alan reviews and summarizes the application of the Hawkes process in the financial field \cite{28hawkes2018hawkes}. Hansen et al. show powerful expressive ability in describing the neural excitation process of Hawkes process in neuroscience with the Lasso (Least absolute shrinkage and selection operator) method \cite{29hansen2015lasso}.

\subsection{Neural Hawkes Process}

Du et al. present the RMTPP models \cite{12zhou2013learning}, which can learn the history effect through RNN, including history event type and time-stamp. This model abandons the restrictions of the Hawkes process and other point process models firstly and achieves improvements than the traditional Hawkes process. Xiao et al. \cite{15xiao2017modeling} utilize two RNN to model the event sequence, one of them is used to model the background intensity and the other is used to model the impact of historical events. Mei and Eisner \cite{14mei2017neural} propose a new LSTM, i.e., the continuous-time LSTM, whose state can decay with time, and based on this LSTM, they come up with the neural Hawkes process to model the asynchronous event sequence.

With the mature self-attention mechanism, the self-attention-based neural Hawkes processes are proposed, the self-attention Hawkes process is the first model which utilizes the self-attention mechanism \cite{23zhang2020self}. Based on the success of the transformer, Zuo et al. utilize the encoder structure in the transformer to get the hidden representation of sequences data, and then convert it to the continuous conditional intensity functions \cite{24DBLP:conf/icml/ZuoJLZZ20}. In recent researches about the neural Hawkes process, the self-attention Hawkes process and transformer Hawkes process achieve great success, which are on the foundation of self-attention mechanism, thus we also focus on the self-attention in our proposed TAA-THP.

\subsection{Transformer models}

\begin{figure}[!htbp]
	\centering
	\includegraphics[scale=1]{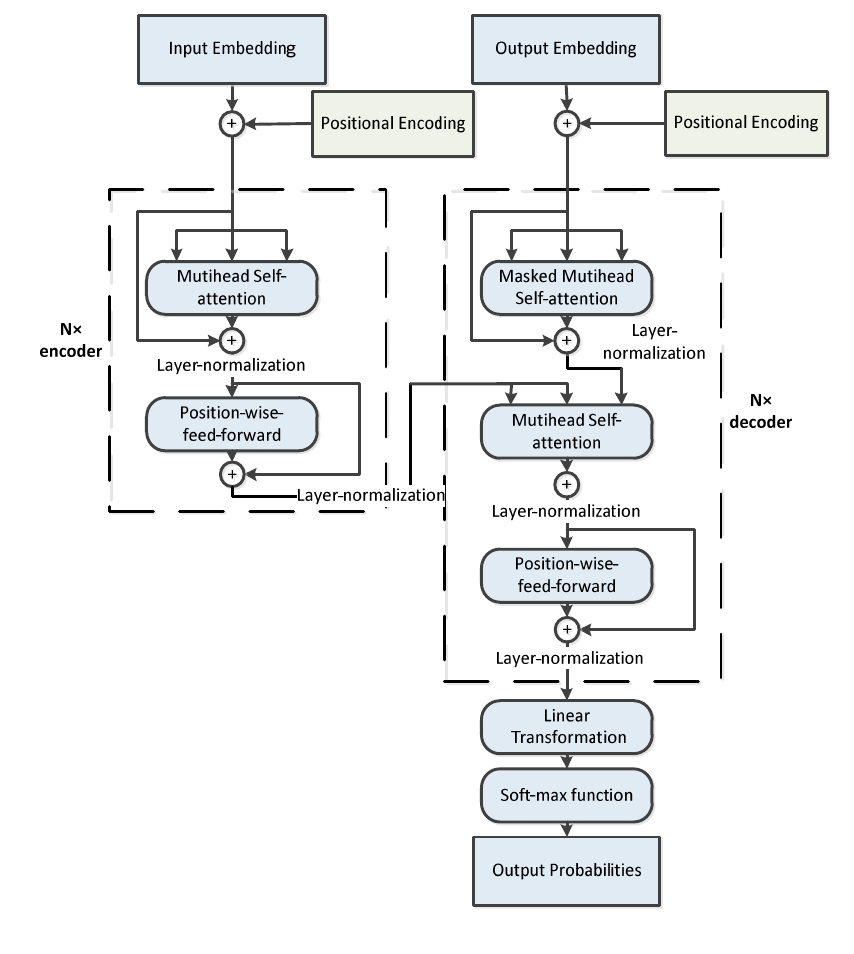}
	\caption{The architecture of transformer model}
	\label{fig1}
\end{figure}

In 2017, Vaswani et al. propose the transformer model \cite{19vaswani2017attention}, which makes full use of self-attention mechanism \cite{18DBLP:journals/corr/BahdanauCB14}, and achieves great success in sequence learning. The architecture of transformer is shown as Fig.\ref{fig1}.

The transformer model is made up of multiple encoder and decoder modules and there are the only self-attention mechanism and position-wise-feed-forward structure without using the recurrence neural network architecture. Meanwhile, recent researches shows that recurrent learning may be more important beyond imagination in sequential learning, thus, Dehghani et al. propose the universal transformer \cite{30DBLP:conf/iclr/DehghaniGVUK19}, which combine the recurrent learning and self-attention mechanism, moreover, in order to better allocate model computing resources, the Adaptive Computation Time (ACT) mechanism \cite{31graves2016adaptive} is introduced into models, then, this model achieves better results than transformers and also achieves Turing completeness. Dai et al. come up with a new transformer called Transformer-XL \cite{25dai2019transformer}, which consists of a segment-level recurrence mechanism and a novel positional encoding scheme, not only does the model perform better, the learning speed is also faster than previous models.

\section{Proposed Model}

In this section, we are going to introduce the details of the temporal attention augmented transformer Hawkes process, including its model structure and corresponding continuous conditional intensity function. We list the notations used in this paper, which is shown in Table 1.

\begin{table}[!htbp]
	\centering
	\caption{Nomenclature}
	\label{tb1}
	\begin{tabular}{ccc}
		\hline	
Symbols	& Description & Size\\ \hline
$S_e$	& The dataset of sequences & /  \\
$s_n$	& The $n$-th sequence & / \\
$I_n$	& The length of the $n$-th sequence & $\mathbb{N}^+$ \\
$N$		& The total number of sequences& $\mathbb{N}^+$ \\
$C$		& The total number of type of events in sequences & $\mathbb{N}^+$ \\
$t_i$		& The time stamp of \textit{i}-th event & $\mathbb{R}$ \\
$c_i$		& The event type of \textit{i}-th event & $\mathbb{N}^+$ \\
$D$		& The model dimension of transformer & $\mathbb{N}^+$ \\
$D_H$		& The dimension of position-wise-feed-forward part of transformer & $\mathbb{N}^+$ \\
$\bm{x}$		& The temporal position encoding & $\mathbb{R}^D$ \\
$\bm{E}$		& The embedding matrix of event type & $\mathbb{R}^{D\times C}$ \\
$\bm{c_i}$		& one-hot encoding vector of each event & $\mathbb{R}^C$ \\
$L$		& The number of multi-head attention & $\mathbb{R}$ \\
$\left\{ {{\mathbf{A}}_l } \right\}_{l = 1}^L $		& The attention variable of multi-head attention & ${\mathbb{R}}^{D_V}$ \\
$D_K$		& The dimension of query and key vector & $\mathbb{N}^+$ \\
$D_V$		& The dimension of value vector & $\mathbb{N}^+$ \\
$\bm{S}$		& State & $\mathbb{R}^{D}$ \\
$\left\{ {{\bm{Q}}_l } \right\}_{l = 1}^L $		& The query variable of multi-head attention & $
\mathbb{R}^{I_n  \times D_K } $ \\ 
$\left\{ {{\bm{K}}_l } \right\}_{l = 1}^L $		& The key variable of multi-head attention & $
\mathbb{R}^{I_n  \times D_K } $ \\ 
$\left\{ {{\bm{V}}_l } \right\}_{l = 1}^L $		& The value variable of multi-head attention & $
\mathbb{R}^{I_n  \times D_V } $ \\  
$\left\{ {{\bm{W}}_Q^l } \right\}_{l = 1}^L$ & The query matrix of multi-head attention & $\mathbb{R}^{D \times D_K } $ \\
$\left\{ {{\bm{W}}_K^l } \right\}_{l = 1}^L$		& The key matrix of multi-head attention & $\mathbb{R}^{D \times D_K } $ \\
$\left\{ {{\bm{W}}_V^l } \right\}_{l = 1}^L$		& The value matrix of multi-head attention & $\mathbb{R}^{D \times D_V } $ \\
$\left\{ {{\bm{W}}_{Tem}^l } \right\}_{l = 1}^L$		& The temporal attention variable of multi-head attention &  $\mathbb{R}^{D \times D_K } $\\
$\left\{ {{\bm{b}}_{lq} } \right\}_{l = 1}^L $	& The bias of dot-product attention & $\mathbb{R}^{D_K } $  \\
$\left\{ {{\bm{b}}_{lq} } \right\}_{l = 1}^L $	& The bias of temporal attention & $\mathbb{R}^{D_K } $  \\
$\bm{W}_multi$		& The aggregation matrix of multi-head attention & $\mathbb{R}^{L{D_V}\times D } $ \\
$\left\{ {{\bm{W}}_i^{FC} ,{\bm{b}}_i^{FC} } \right\}_{i = 1}^4$
		& The parameters of fully connected neural network & / \\
$\bm{H}$		& The hidden representation of sequence &   $\mathbb{R}^{I_n\times D}$ \\
$\bm{h}(t_i)$	& The hidden representation of \textit{i}-th event & $\mathbb{R}^{D}$ \\
$\mathcal{H}_t$	 & The previous history at time \textit{t} &  \\
$b_c$		& The background intensity of event type \textit{c} & $\mathbb{R}$ \\
${\bm{w}}_{\alpha _c } $	& The weight of continuous parameter of conditional intensity function of event type & $\mathbb{R}^{1\times D}$ \\
${\bm{w}}_c^T $& Historical weight parameter of conditional intensity function of event  type & $\mathbb{R}^{1\times D}$ \\
$\bm{W}_{time}$		& The prediction parameter of time-stamp & $\mathbb{R}^{1\times D}$ \\
$\bm{W}_{type}$		&The prediction parameter of event type  & $\mathbb{R}^{C\times D}$ \\ \hline
	\end{tabular}
\end{table}

\subsection{Temporal Attention Augmented Transformer Hawkes Process}

\begin{figure}[!htbp]
	\centering
	\includegraphics[scale=1]{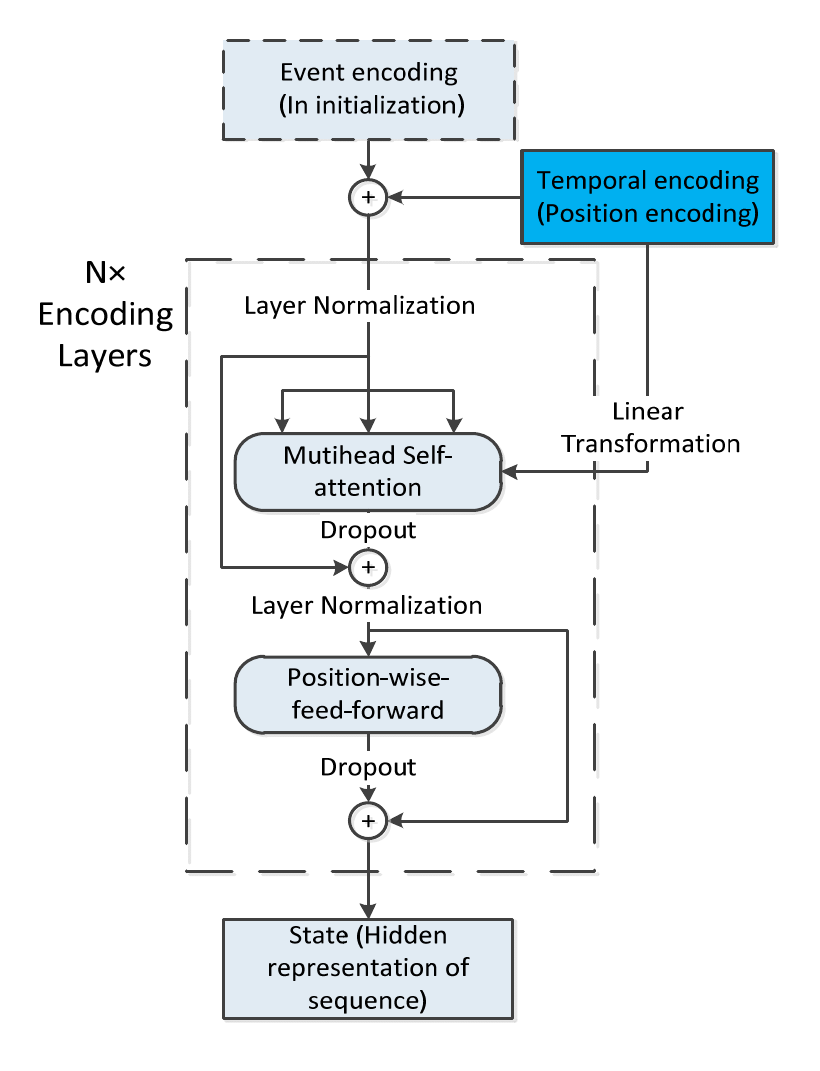}
	\caption{The schematic diagram of Temporal Attention Augmented Transformer Hawkes Process (TAA-THP), in TAA-THP, we modify the traditional attention mechanism and highlight the use of temporal encoding information of event sequence.}
	\label{fig2}
\end{figure}

Generally speaking, transformer-based Hawkes process model utilizes the encoder structure of different kinds of the transformer to get the hidden representation of event sequence. Assuming there are  sequences in the dataset $S_e$ , and the $n$-th sequence is $s_n  = \{ t_i ,c_i \} _{i = 1}^{I_n } $ , whose length is $I_n$ . Each pair in $S_n$ is composed with two parts, e.g., $t_i$ and $c_i$ . Among them, $c_i$ is the type of event that occurred and $t_i$ is the corresponding time-stamp. Following point process theory, time-stamp is the event occurring instant in tandem along the time axis, which is consistent with the position encoding of the transformer-based model for asynchronous sequences. Thus, we can make use of this general method to encode the timestamp as a positional encoding, similar to other transformer-based models \cite{19vaswani2017attention}, and \cite{24DBLP:conf/icml/ZuoJLZZ20}, which is shown in Eq.\ref{eq2}:

\begin{equation}
	\label{eq2}
[{\bm{x}}(t_i )]_j  = \left\{ {\begin{array}{*{20}c}
		{\cos \left( {t_i /10000^{\frac{{j - 1}}{D}} } \right),{\rm{if}}\:j\:{\rm{is}}\:{\rm{odd}},}  \\
		{\sin \left( {t_i /10000^{\frac{j}{D}} } \right),{\rm{if}}\:j\:{\rm{is}}\:{\rm{even}}.}  \\
\end{array}} \right.	
\end{equation}

where $\bm{x}$ is the position encoding for transformer model, throughout the latter part of the paper, we will call it as temporal encoding, and $D$ is the dimension of transformer model. For the event encoding, we utilize one-hot encoding vector of each event ${\mathbf{c}}_i  \in \mathbb{R}^C $, and embedding matrix ${\mathbf{E}} \in \mathbb{R}^{D \times C}$ to get it. In this way, we can get ${\bm{X}}^T$ and $({\bm{EC}}_n )^T $, which are the corresponding temporal encoding and event encoding of the sequence, where ${\bm{X}} = \{ {\bm{x}}(t_1 ),{\bm{x}}(t_2 ),...,{\bm{x}}(t_{I_n } )\}  \in \mathbb{R}^{D \times I_n }$  and $
{\bm{C}}_n  = [{\bm{c}}_1 ,{\bm{c}}_2 ,...,{\bm{c}}_{I_n } ] \in \mathbb{R}^{C \times I_n } $ . So as to obtain the hidden representation of event sequence, we input the ${\bm{X}}^T$ and $({\bm{EC}}_n )^T $ to the model we proposed, whose schematic diagram is shown in Fig.\ref{fig2}.

As shown in Fig.\ref{fig2}, in initial, event and temporal encoding are inputted to the model, and the input of each encoding layer is the output of the previous layer plus temporal encoding. Which means in the first layer of TAA-THP, we can calculate . Meanwhile, we modify the multi-head self-attention and add temporal encoding as an additional auxiliary input. The tricks, such as layer normalization, residual connection, and dropout, are also involved in TAA-THP, to utilize the information of temporal encoding   iteratively, the traditional dot-product multi-head self-attention in encoding layers are modified by us, the new dot-product multi-head self-attention with temporal augmented attention is written as follows:

\begin{equation}
	\label{eq3}
	{\bm{A}}_l  = Softmax\left[ {mask\left( {\frac{{\left( {{\bm{Q}}_l  + {\bm{b}}_{lq} } \right){\bm{K}}_l^T  + \left( {{\bm{Q}}_l  + {\bm{b}}_{lt} } \right)\left( {{\bm{X}}^T {\bm{W}}_{Tem}^l } \right)^T }}
			{{\sqrt {D_K } }}} \right)} \right]{\bm{V}}_l 	
\end{equation}

In Eq.3, ${\bm{b}}_{lq}$ and ${\bm{b}}_{lq }$ are the bias vectors of query matrix , these biases will help the model to calculate the attention more flexible. The term  $\left( {{\mathbf{Q}}_l  + {\mathbf{b}}_{lq} } \right){\mathbf{K}}_l^T $ is the modified dot-product attention operation, which just adds a bias term in traditional dot-product attention operation. We focus on another term, i.e., $\left( {{\mathbf{Q}}_l  + {\mathbf{b}}_{lt} } \right)\left( {{\mathbf{X}}^T {\mathbf{W}}_{Tem}^l } \right)^T$ , this term re-introduces the temporal information into the dot-product, rather than just add the temporal encoding to $\bm{S}$ . In this term, ${\bm{X}}^T {\bm{W}}_{Tem}^l$  is the linear transformation of temporal encoding ${\bm{X}}^T$ ,${\bm{W}}_{Tem}^l  \in \mathbb{R}^{D \times D_K } $ is the linear transformation matrix of the $l$-th head attention, and for the $l$-th query, key and value matrix are obtained from Eq.\ref{eq4}:

\begin{equation}
	\label{eq4}
	{\bm{Q}}_l  = {\bm{SW}}_Q^l ,{\bm{K}}_l  = {\bm{SW}}_K^l ,{\bm{V}}_l  = {\bm{SW}}_V^l 
\end{equation}

where $\bm{S}$ , as described in following Algorithm 1, is the input of each encoding layer.${\bm{W}}_Q^l ,{\bm{W}}_K^l  \in \mathbb{R}^{D \times D_K }$  and ${\bm{W}}_V  \in \mathbb{R}^{D \times D_V }$ are a linear transformation of $\bm{S}$ . And in general, for our proposed transformer model TAA-THP, we impose the constraint $D_K  = D_V $, which is similar to the usual conventions. Meanwhile, in order to shield the impact of future events on current events (one-way characteristics of time) we utilize the masked self-attention mechanism similar to \cite{18DBLP:journals/corr/BahdanauCB14}, which set the elements above the main diagonal of the matrix to be negative infinity, the function $mask( \cdot )$ is used to ensure that future events in the matrix will not affect the attention weights of current events.

By means of Eq.\ref{eq3} and Eq.\ref{eq4}, we can obtain , which is the  $l$-th attention matrix. To improve the expressive ability of the model, most of the existing attention mechanism models use multi-head attention. Assuming there are $L$ self-attention heads, then, we can get $L$ outputs ${\bm{A}}_1 ,{\bm{A}}_2 ,...,{\bm{A}}_L $  based on Eq.\ref{eq3} and Eq.\ref{eq4}, the corresponding parameters are $\left\{ {{\bm{W}}_Q^l ,{\bm{W}}_K^l ,{\bm{W}}_{Tem}^l ,{\bm{W}}_V^l } \right\}_{l = 1}^L $ . The formula for combining multi-head attention into one output is described as Eq.\ref{eq5}:

\begin{equation}
	\label{eq5}
{\mathbf{A}} = \left[ {{\mathbf{A}}_1 ,{\mathbf{A}}_2 ,...,{\mathbf{A}}_L } \right]{\mathbf{W}}_{multi} 
\end{equation}

where ${\bm{W}}_{multi}  \in \mathbb{R}^{LD_V  \times D}$ is the aggregation matrix. In the position-wise-feed-forward part of each encoding layer, a convolutional neural network (CNN) module is added between the two fully connected layers, the CNN implement the following calculation:

\begin{equation}
	\label{eq6}
{\bm{S}} = {\rm{CNN}}\left( {{\rm{ReLU}}({\bm{AW}}_1^{FC}  + {\bm{b}}_1 )} \right){\bm{W}}_2^{FC}  + {\bm{b}}_2	
\end{equation}

The CNN module consists of a one-dimension convolutional layer, a nonlinear activation function (ReLU function) and a one-dimension max-pooling layer. Based on the thumb rule and the cross-validation results, we set the convolution kernel have 1 input channel, 4 output channels, the size of the convolution kernel is 3, the stride is 2, and the padding is 0, and the size and stride of max-pooling layer are 2.

Our aim for incorporating the CNN module is to improve the local perception of the model so that the model can better learn the dependencies between events. In summary, the algorithm for calculating the implicit representation of the event sequence through asynchronous event sequence data is shown in Algorithm \ref{alg1}.

\begin{algorithm}
	\renewcommand{\algorithmicrequire}{\textbf{Input:}}
	\renewcommand{\algorithmicensure}{\textbf{Output:}}
	\caption{Temporal Attention Augmented Transformer Hawkes Process (TAA-THP)}
	\label{alg1}
	\begin{algorithmic}[2]
		\REQUIRE The number of encoding layers: $n$, event-type encoding $({\bm{EC}}_n )^T$
		. Temporal encoding (position encoding) ${\bm{X}}^T$ .
		\ENSURE Hidden representation (State) $	{\mathbf{H}} \in \mathbb{R}^{D \times I_N } $of event sequence  
		\STATE  Initialize state $ {\bm{S}} \leftarrow ({\bm{EC}}_n )^T $ .
		\STATE \textbf{for} $i$ \textbf{in} $n$ ,
		\STATE ${\bm{S}} \leftarrow {\bm{S}} + {{ }}{\bm{X}}^T $
		,
		\STATE ${\bm{S}} \leftarrow output\;of\;{\rm{the}}\;i{\rm{ - th }}\;{\rm{encoding\;layer}}({\bm{S}},{\bm{X}}^T )$	
		\STATE \textbf{return} ${\bm{H}} \leftarrow {\bm{S}}$
	\end{algorithmic}
\end{algorithm}

And inspired by \cite{24DBLP:conf/icml/ZuoJLZZ20} and \cite{32wang2019language}, using Algorithm \ref{alg1}, we will get the output $\bm{H}$ , and $\bm{H}$  will be utilized as the input of a postprocessing part, which consists of a fully connected layer, RNN layer, and another fully connected layer. The structure of the postprocessing part is summed up in Eq. \ref{eq7}:

\begin{equation}
	\label{eq7}
	{\bm{H}} = {\rm{RNN}}\left( {{\rm{ReLU}}({\bm{HW}}_3^{FC}  + {\bm{b}}_3 )} \right){\mathbf{W}}_4^{FC}  + {\bm{b}}_4 	
\end{equation}

Based on cross-validation and other related researches, we set the network dimension of RNN to zero (which means no postprocessing part) or 64. The types of RNN networks that can be used include but are not limited to LSTM and GRU, after $\bm{H}$ passes through this part, the dimension of $\bm{H}$ remains unchanged. Whether the post-processing part in the proposed TAA-THP can improve the performance of the model depends on the characteristics of the dataset.

\subsection{Conditional Intensity Function}

The relationship between the probability of occurrence and occurring time of each type of event can be described by its corresponding conditional intensity function, after getting the hidden representation of sequence, we can utilize it to calculate the conditional intensity function $\lambda _c (t\left| {\mathcal{H}_t } \right.)$, where $\mathcal{H}_t  = (t_j ,c_j ):t_j  < t$, which is the history before time  , we adopt the similar approach in [24] to calculate the conditional intensity function, which is shown as Eq.\ref{eq8}:

\begin{equation}
	\label{eq8}
	\lambda _c (t\left| {\mathcal{H}_t } \right.) = f(b_c  + \alpha _c \frac{{t - t_i }}
	{{t_i }} + {\bm{w}}_c^T {\bm{h}}(t_i ))	
\end{equation}

where $t_i$  is the time-stamp of the last event which its occurrence time is the closest to $t$, and $\bm{h}_t$ is the hidden representation corresponding to this event. The overall background intensity is $b_c  + \alpha_c {\bm{h}}(t_i )\frac{{t - t_i }}
{{t_i }}$ , here $t \in [t_i ,t_{i + 1} )$ , $b_c$ is the constant background intensity,  $\alpha_c$ is the coefficient of time-dependent background intensity. In \cite{24DBLP:conf/icml/ZuoJLZZ20}, the $\alpha_c$  is fixed to -0.1, which indicates that when no event occurs, the conditional intensity function will inevitably decay over time. This assumption does not completely conform to reality and limits the expressive ability of the model. In TAA-THP, to make up for this drawback, we let $\alpha_c  = {\bm{w}}_{\alpha_c } {\bm{h}}(t_i )$, i.e., the  background intensity  change with the past history, which improves the model's ability to fit the conditional intensity function. Compared with the traditional Hawkes process model, the time-dependent background intensity can better fit the basic probability of event occurrence. $f(x) = \frac{1} {\beta }\log (1 + e^{\beta x} ) $ is softplus function, and $\beta$ is the softness parameter of softplus function. Softplus function is an improved version of the ReLU function, compared with ReLU, this nonlinear function is smoother, and the activity of most neurons is guaranteed, which makes the model learning ability stronger. 

Finally, the overall conditional intensity function is the sum of the conditional intensity functions of all types of events, as shown in Eq.\ref{eq9}:  

\begin{equation}
	\label{eq9}
\lambda (t\left| {\mathcal{H}_t } \right.){\rm{  =  }}\sum\limits_{c = 1}^C {\lambda _c (t\left| {\mathcal{H}_t } \right.)} 
\end{equation}

\section{Prediction Task and Model Training}

In general, the objective function of the point process model is constructed from the likelihood function \cite{7daley2007introduction}, more specifically, for a sequence $s_n  = \{ t_i ,c_i \} _{i = 1}^{I_n }$ , the log-likelihood function of $s_n$ is shown as Eq.\ref{eq10}:

\begin{equation}
	\label{eq10}
L(s_n ) = \sum\limits_{i = 1}^{I_n } {\log } \,\lambda (\left. {t_i } \right|\mathcal{H}_i ) - \int_{t_1 }^{t_{I_n } } {\lambda (\left. t \right|\mathcal{H}_t )} dt
\end{equation}

and assuming there are $N$ sequences in the dataset, then the model parameters can be solved by the maximum log-likelihood principle:

\[
\max \sum\nolimits_{n = 1}^N {L(s_n )} 
\]

However, let $\Lambda  = \int_{t_1 }^{t_{I_n } } {\lambda \left( {\left. t \right|\mathcal{H}_t } \right)} dt
$ , due to $\lambda (\left. t \right|\mathcal{H}_t )$  is derived from the neural network, the closed-form of  is hard to get, and we have two approximate methods to solve it, Monte Carlo sampling method \cite{33robert2013monte} and the numerical analysis method \cite{34stoer2013introduction}. The Monte Carlo sampling method \cite{33robert2013monte}, the approximate value of $\Lambda$ is shown in Eq.\ref{eq11}:

\begin{equation}
	\label{eq11}
\hat \Lambda _{MC}  = \sum\limits_{i = 2}^L {\left( {t_i  - t_{i - 1} } \right)} (\frac{1}
{M}\sum\limits_{m = 1}^M {\lambda (u_m )} )
\end{equation}

where $u_m$ is sampled from the uniform distribution $U(t_{i - 1} ,t_i )$ . It is worth noting that $\hat \Lambda _{MC} $ calculated by this method is an unbiased estimate of $\Lambda$. And for the numerical analysis method \cite{34stoer2013introduction}, based on the trapezoidal rule, we can get the estimate of $\Lambda$ as shown in Eq.\ref{eq12}:

\begin{equation}
	\label{eq12}
\hat \Lambda _{NA}  = \sum\limits_{i = 2}^{I_n } {\frac{{t_i  - t_{i - 1} }}
	{2}} (\lambda (\left. {t_i } \right|\mathcal{H}_i ) + \lambda (\left. {t_{i - 1} } \right|\mathcal{H}_{i - 1} ))
\end{equation}

The computational complexity of the numerical analysis method is significantly less than the Monte Carlo sampling method, but the error is larger. It is a reasonable alternative when the accuracy requirements are not high.
Below we give a more scrutinizing discussion. Recall that after obtaining the conditional intensity function of each type of event through Eq.\ref{eq8}, we can predict the occurrence instant of events in the future. Traditionally \cite{7daley2007introduction}, the dominant paradigm to calculate the conditional intensity function can be gotten from Eq.\ref{eq13}:

\begin{equation}
	\label{eq13}
\begin{array}{l}
	p(\left. t \right|{\cal H}_t ) = \lambda (\left. t \right|{\cal H}_t )\exp t( - \int_{t_i }^t {\lambda (\left. s \right|{\cal H}_t )} ds) \\ 
	\hat t_{i + 1}  = \int_{t_i }^\infty  {t\cdot p(\left. t \right|{\cal H}_t )dt}  \\ 
	\hat c_{i + 1}  = \mathop {\arg \max }\limits_c \frac{{\lambda _c (\hat t_{i + 1} |{\cal H}_{i + 1} )}}{{\lambda (\hat t_{i + 1} |{\cal H}_{i + 1} )}} \\ 
\end{array}
\end{equation}

However, this paradigms are flexible, but highly intractable, which have  some limitations, for instance, $\lambda (\left. t \right|{\cal H}_t )$ is derived from the neural network, which is hard to get the closed-form solution of $\int {t\cdot p(\left. t \right|{\cal H}_t )dt} $ , moreover, we need to calculate the improper integral $\int_{t_i }^\infty  {t\cdot p(\left. t \right|{\cal H}_t )dt} $ based on Monte Carlo sampling \cite{33robert2013monte}, generally speaking, the upper limit of integration can be replaced by a large enough number when we take advantage of  Monte Carlo sampling from the uniform distribution to approximately calculate the integral, but this still has a larger error compared with the upper limit of integration being infinite, Monte Carlo sampling under the exponential distribution can take the upper limit of integration to infinity, while high computational complexity limits the utility of this approach. Therefore, we desire to use the powerful fitting ability of the neural network and the extracted hidden representation from the sequence of events,  to predict the type and time of events that may occur in the future. The prediction formula is presented as Eq.\ref{eq14}:

\begin{equation}
	\label{eq14}
\begin{array}{l}
	\hat t_{i + 1}  = {\bm{W}}_{time} {\bf{h}}(t_i ) \\ 
	\widehat{\bf{p}}_{i + 1}  = Softmax({\bm{W}}_{type} {\bf{h}}(t_i )) \\ 
	\hat c_{i + 1}  = \mathop {\arg \max }\limits_c {\bf{\hat p}}_{i + 1} (c) \\ 
\end{array}
\end{equation}

In addition, the prediction parameter, ${\bm{W}}_{time}$  and ${\bm{W}}_{type}$ , should also be optimized, we define the prediction loss of occurring time of each type of event and type of events as Eq.\ref{eq15} and Eq.\ref{eq16}:

\begin{equation}
	\label{eq15}
	L_{time} (s_n ) = \sum\nolimits_{i = 2}^{I_n } {(t_i  - \hat t_i )^2 } 
\end{equation}

\begin{equation}
	\label{eq16}
	L_{type} (s_n ) = \sum\nolimits_{i = 2}^{I_n } { - {\bf{c}}} _i^T \log (\widehat{\bf{p}}_i )
\end{equation}

Thus, the overall objective function of TAA-THP is shown as Eq.\ref{eq17}:

\begin{equation}
	\label{eq17}
\min \sum\limits_{n = 1}^N { - L(s_n )}  + \alpha _{type} L_{type} (s_n ) + \alpha _{time} L_{time} (s_n )
\end{equation}

where $\alpha _{type} $ and $\alpha _{time}$ are hyperparameters, which will help keep the stability of model
training, the objective function in Eq. 17, can be optimized by stochastic gradient descent algorithm
including but not limited to momentum method and ADAM (Adaptive moment estimation)\cite{35kingma2014adam},
we use ADAM optimizer with default hyperparameter i.e.,
we set learning rate=0.0001, betas are 0.9 and 0.999, eps =$10^8$, weight\_decay = 0.
\section{Experiments}

In this section, we are going to compare the model performance of TAA-THP and numerous baselines on a wide range of synthetic and real-life datasets, and figure out the effect of the introduction of temporal augmented attention. Firstly, we introduce the details of datasets and baselines, then, we compare the model performance on the foundation of experimental results, at last, we explore the effect of the presence or absence of temporal augmented attention on the performance of the model in the ablation study.

\subsection{Datasets}

In this subsection, we make use of two artificial datasets, and four real-world datasets of event sequence to conduct the experiments, Table 2 introduces the characteristics of each dataset:

\begin{table}[!htbp]
	\centering
	\label{tb2}
	\caption{Characteristics of datasets used in experiments.}
	\begin{tabular}{cccccc}
		\hline
		\multirow{2}{*}{Dataset} & \multirow{2}{*}{C} & \multicolumn{3}{c}{Sequence   Length} & \multirow{2}{*}{Events} \\ \cline{3-5}
		&                    & Min        & Aver.       & Max        &                         \\ \hline
		Synthetic                & 5                  & 20         & 60          & 100        & 602,697                 \\
		NeuralHawkes             & 5                  & 20         & 60          & 100        & 602,984                 \\
		Retweets                 & 3                  & 50         & 109         & 264        & 2,173,533               \\
		MIMIC-II                 & 75                 & 2          & 4           & 33         & 2,419                   \\
		StackOverflow            & 22                 & 41         & 72          & 736        & 480,413                 \\
		Financial                & 2                  & 829        & 2074        & 3319       & 414,800    \\ \hline            
	\end{tabular}
\end{table}

\textbf{Synthetic and NeuralHawkes} \cite{14mei2017neural}: Mei generates two sets of artificial sequences based on the thinning algorithm, including Hawkes process model and continues-time LSTM neural Hawkes process model, the parameters are randomly sampled.

\textbf{Retweets} \cite{6zhao2015seismic}:This dataset contains a large number of sequences of tweets, every sequence includes an original tweet (the original content is some user post) and its following retweets. The time and label of the user of each retweet are recorded in the sequence, and labels are divided into three classes based on the number of their followers: “small”, “medium”, and “large”.

\textbf{StackOverflow} \cite{36leskovec2014snap}: StackOverflow is one of the most famous programming Q\&A websites. The website rewards users with various badges to encourage them to take part in community activities. It is worth noting that the same badge can be donated to the same user, and this dataset includes lots of users’ reward histories in the range of two years, each user’s history is treated as a sequence, and each event in sequences represents the acquisition of badge.

\textbf{Electrical Medical Records} \cite{2johnson2016mimic}: MIMIC-II dataset includes patients’ records of visitation to a hospital’s ICU during seven years. Each patient’s record is treated as an independent sequence, and each event contains the corresponding time-stamp and diagnosis of the patient.

\textbf{Financial Transactions} \cite{12zhou2013learning}: The financial dataset involves a lot of short-term transaction history of a stock in one day. The operation of each transaction is recorded as the event, and this dataset only has two kinds of events: “buy” and “sell”, and the unit of the time-stamp in milliseconds.
Based on the cross-validation results on validation dataset, we obtain the hyper-parameters of our model on different datasets, which are shown as Table 3.

\begin{table}[!htbp]
	\centering
	\label{tb3}
	\caption{The hyper-parameters of Temporal Attention Augmented Transformer Hawkes Process}
	\begin{tabular}{ccccccccc}
		\hline
		Dataset       & Batch Size & $D$   &  $D_H$   &   $D_K=D_V$  & Heads of attention & Layers of model &  $D_{\rm{RNN}}$  & Dropout \\ \hline
		Synthetic     & 16         & 64  & 256 & 64  & 3                  & 3               & 64 & 0.1     \\
		NeuralHawkes  & 16         & 64  & 256 & 64  & 3                  & 3               & 64 & 0.1     \\
		Retweets      & 16         & 64  & 256 & 64  & 3                  & 3               & 64 & 0.1     \\
		MIMIC-II      & 1          & 128 & 256 & 256 & 5                  & 5               & 0  & 0.1     \\
		StackOverflow & 4          & 128 & 512 & 256 & 4                  & 4               & 64 & 0.1     \\
		Financial     & 1          & 128 & 512 & 512 & 4                  & 4               & 64 & 0.1     \\ \hline
	\end{tabular}
\end{table}

\subsection{Baselines}

\textbf{RMTPP} \cite{12zhou2013learning}: Du et al. come up with a recurrent network point process model, which effectively models the event types and time-stamp in the sequence by embedding history to the vector.

\textbf{NHP} \cite{14mei2017neural}: Mei and Eisner firstly present a novel neural Hawkes process, which based on the continuous-time LSTM, the continuous-time LSTM has decay property of impact of historical events.

\textbf{SAHP} \cite{23zhang2020self}: Zhang et al. firstly come up with the sequence encoding through the self-attention mechanism, which utilizes the position encoding method to encode the time-stamp, and after the implicit representation of the sequence is obtained, the implicit representation of the sequence is used to calculate the model parameters of the Hawkes process.

\textbf{THP} \cite{24DBLP:conf/icml/ZuoJLZZ20}: Base on the performance improvement of the transformer, Zuo et al. propose the transformer Hawkes process, which achieves state-of-the-art performance. We can obtain the following experimental results to verify the performance of THP based on the model and the hyper-parameter they provide.

\subsection{Experimental results and comparison}

In this subsection, in order to validate the performance of TAA-THP, we compare the TAA-THP and baselines’ performance. First of all, we compare the loglikelihood value of TAA-THP and baselines on the test dataset, and for simplicity, we named it Loglike. The Loglike is the most basic model measurement for the point process model, which indicates how well the model fits in the data. In our experimental process, we set $\alpha_time $ and $\alpha_type $ in TAA-THP to zero. The loglike of baselines and TAA-THP on different test datasets are shown in Table 4.

\begin{table}[!htbp]
	\centering
	\caption{The value of log-likelihood function of different models on the test datasets.}
	\begin{tabular}{cccccc}
		\hline
		Datasets      & RMTPP            & NHP   & SAHP             & THP    & TAA-THP \\ \hline
		Synthetic     & \textbackslash{} & -1.33 & 0.520            & 0.834  & \textbf{1.66}    \\
		NeuralHawkes  & \textbackslash{} & -1.02 & 0.241            & 0.966  & \textbf{1.77 }   \\
		Retweets      & -5.99            & -5.06 & -5.85            & -4.69  & \textbf{-1.04}   \\
		StackOverflow & -2.60            & -2.55 & -1.86            & -0.559 & \textbf{-0.545}  \\
		MIMIC-II      & -1.35            & -1.38 & -0.520           & -0.143 & \textbf{-0.111}  \\
		Financial     & -3.89            & -3.60 & \textbackslash{} & -1.388 & \textbf{-1.18}   \\ \hline
	\end{tabular}
\end{table}

As we can find out in Table 4, TAA-THP outperforms existing baselines in terms of the value of the log-likelihood function on all test datasets. This phenomenon proves that TAA-THP can more effectively model the event sequences than existing baselines. However, the acquisition of Loglike can’t provide effective help for the practical application of the model, because loglike is only an abstract measurement, it does not have practical meaning.

Thus, we tend to compare the prediction accuracies in terms of the occurring time of each type of event and type of events for existing baselines and our proposed TAA-THP model, which have very important practical application significance. For instance, in terms of the electronic medical records, based on the patient's medical history, we can predict the type of illness and the occurring time of the next attack, and provide more effective and timely help for the patient's treatment. We can also predict the next possible failure time and type for a large-scale system based on the failure sequence, and perform maintenance and prevention in advance to improve safety and economic benefits in production. 

It is worth noting that, for any sequence of events $s_n$ , assuming the length is $I_n$ , we don’t predict the possible type and time of the first event, in other words, for the sequence $s_n$ , we make n-1 predictions, and we set $\alpha_time=0.01 $ and $\alpha_type=1 $ . The prediction accuracies of event types are shown in Table 5 and Fig. 3.

\begin{table}[]
	\centering
	\caption{Predict accuracies for different models on various datasets.}
	\begin{tabular}{ccccc}
		\hline
		Dataset       & RMTPP & NHP   & THP   & TAA-THP \\ \hline
		StackOverflow & 45.9  & 46.3  & 46.79 & 46.83   \\
		MIMIC-II      & 81.2  & 83.2  & 83.2  & 84.4    \\
		Financial     & 61.95 & 62.20 & 62.23 & 62.43   \\ \hline
	\end{tabular}
\end{table}

\begin{figure}[!htbp]
	\label{fig3}
	\centering
	\subfigure[]{
		\includegraphics[width=4.5cm]{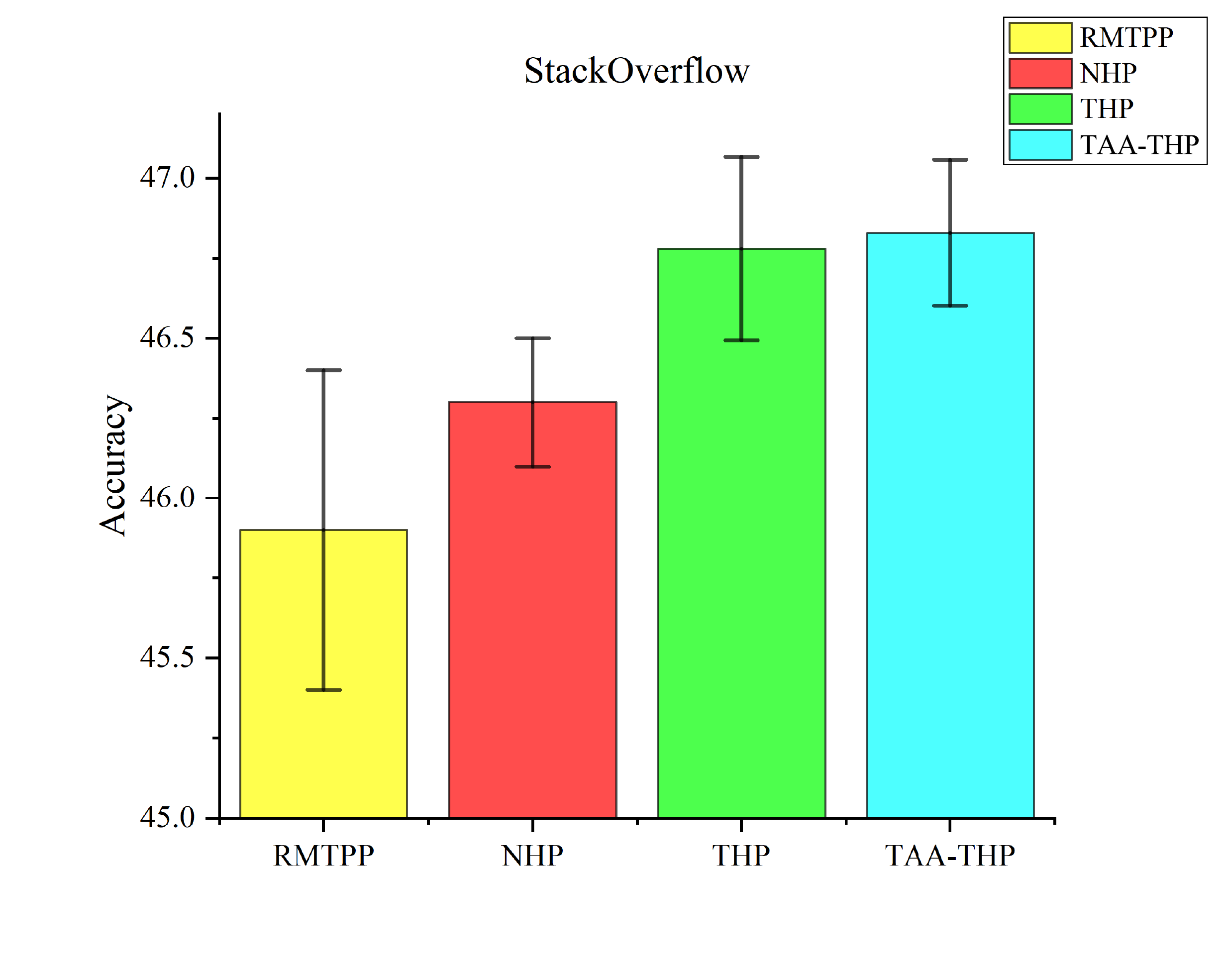}
	}
	\quad
	\subfigure[]{
		\includegraphics[width=4.5cm]{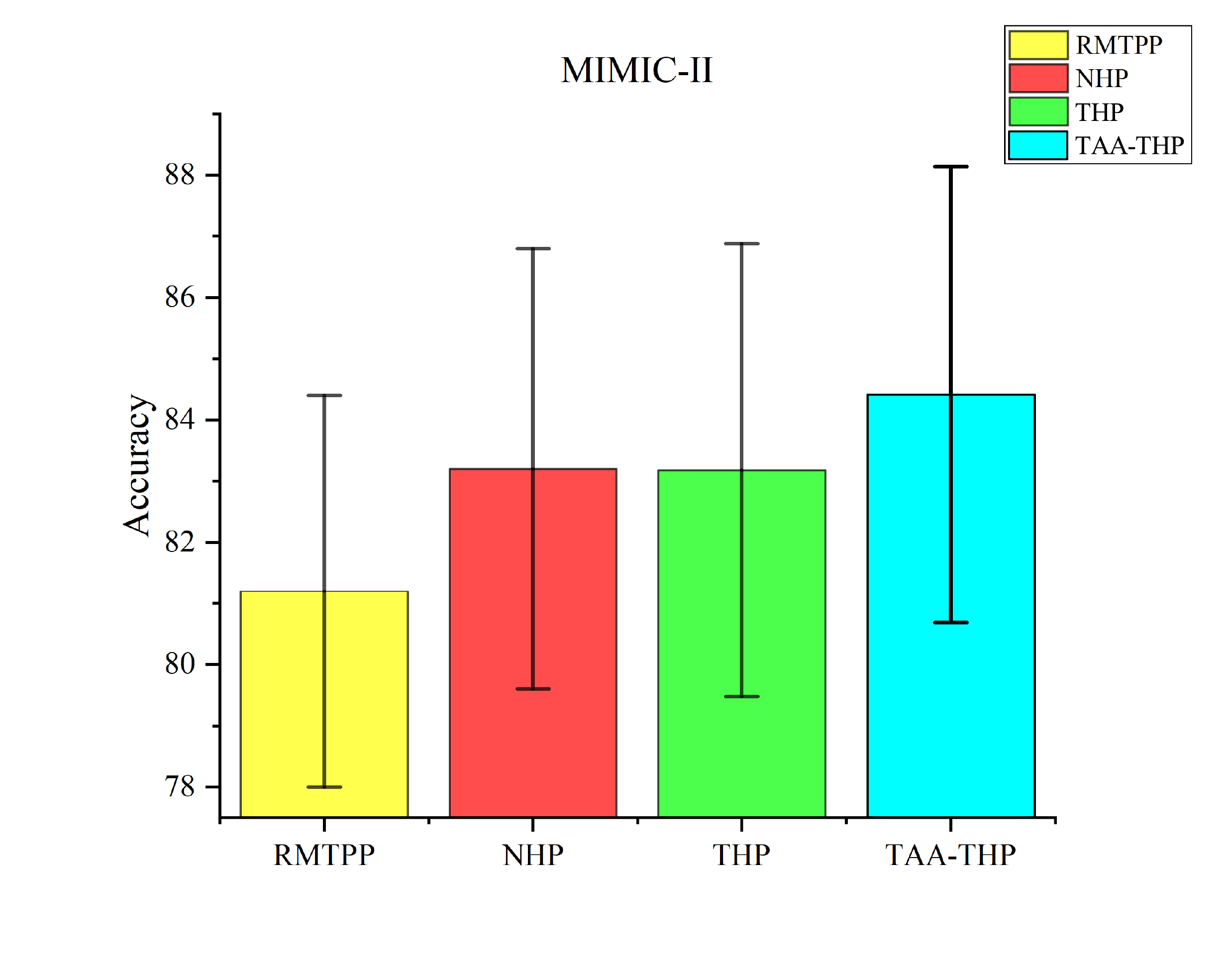}
	}
	\quad
	\subfigure[]{
		\includegraphics[width=4.5cm]{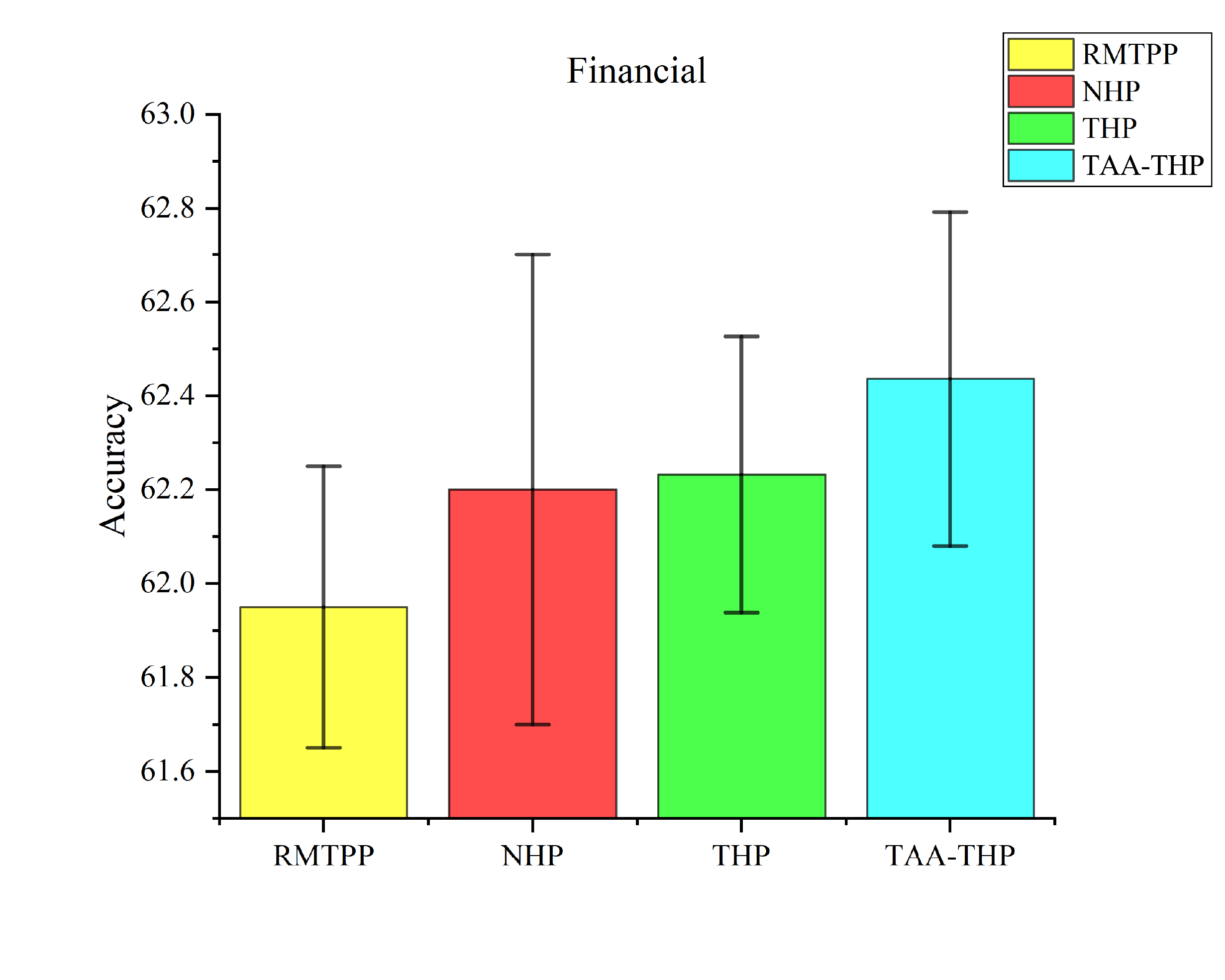}
	}
	\caption{Predictive accuracies for existing baselines and TAA-THP. Based on the five times train-dev-test partition, five experiments are performed on each dataset, and then the mean and standard deviation of different models are obtained.}
\end{figure}

From Fig. 3 and Table 5, on these real-world datasets, we can see that the accuracies of event prediction of our TAA-THP model have been improved compared with the existing baselines. In our point of view, we think these results are caused by the introduction of temporal attention, which decouples the simple stacking of event encoding and temporal position encoding in the traditional dot product attention mechanism. And we will verify this hypothesis on the ablation study in subsection 5.4.
 
However, we can see that the improvement of prediction accuracies is relatively small, however, The experimental results of Loglike have confirmed that TAA-THP can better model the sequence data, and then we will test the prediction ability of occurrences of the events for different models.

As for the prediction error of the occurring time of each type of event, we use RMSE (Root Mean Squared Error) as uniform metric evaluation criteria, RMSE of the existing baselines and TAA-THP are shown in Table 6 and Fig. 4.

\begin{table}[!htbp]
	\centering
	\caption{RMSE of different models on various datasets}
	\begin{tabular}{ccccc}
		\hline
		Dataset       & RMTPP & NHP  & THP     & TAA-THP \\ \hline
		StackOverflow & 9.78  & 9.83 & 4.99    & 3.91    \\
		MIMIC-II      & 6.12  & 6.13 & 0.859   & 0.868   \\
		Financial     & 1.56  & 1.56 & 0.02575 & 0.02570 \\ \hline
	\end{tabular}
\end{table}

\begin{figure}[!htbp]
	\label{fig3}
	\centering
	\subfigure[]{
		\includegraphics[width=4.5cm]{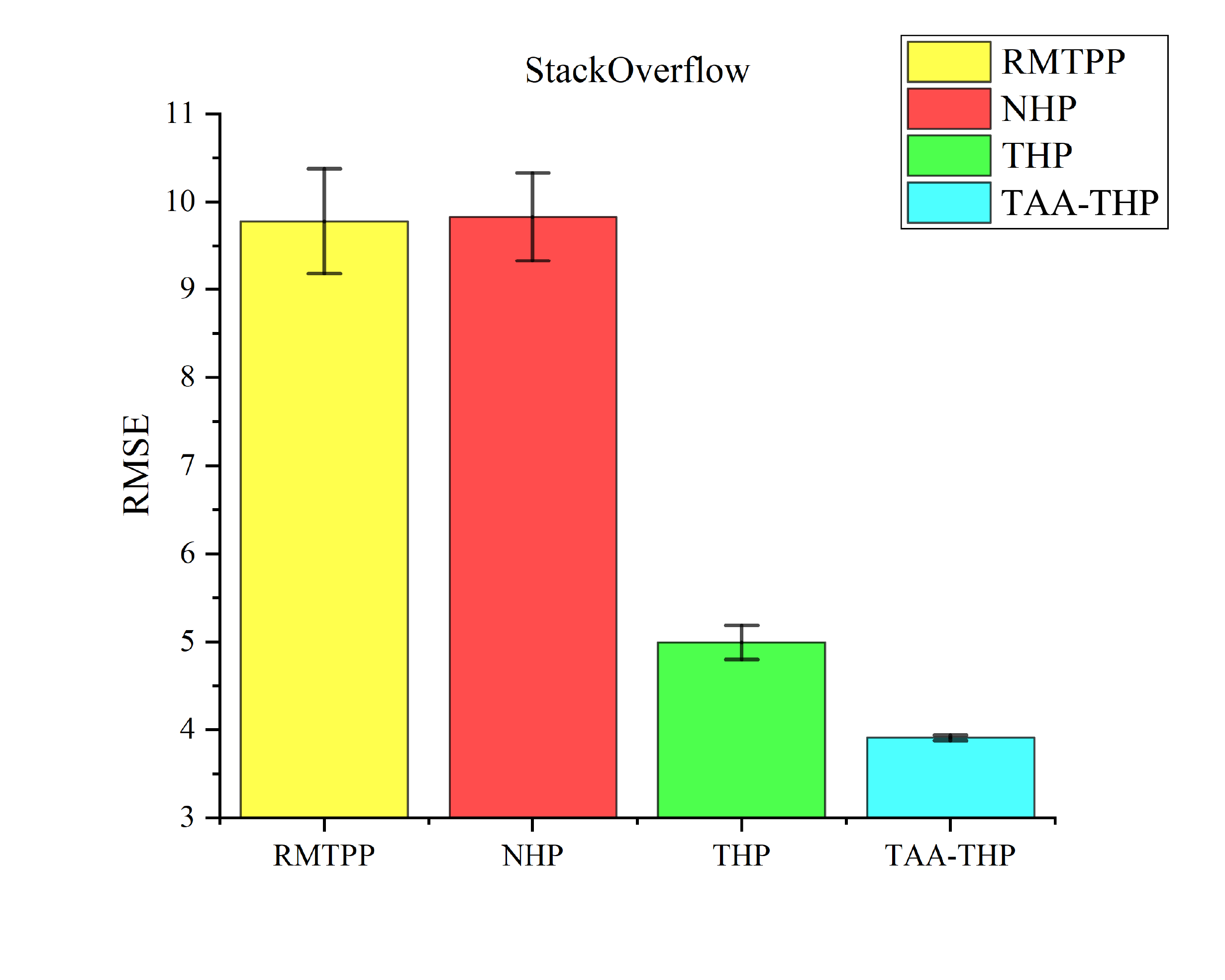}
	}
	\quad
	\subfigure[]{
		\includegraphics[width=4.5cm]{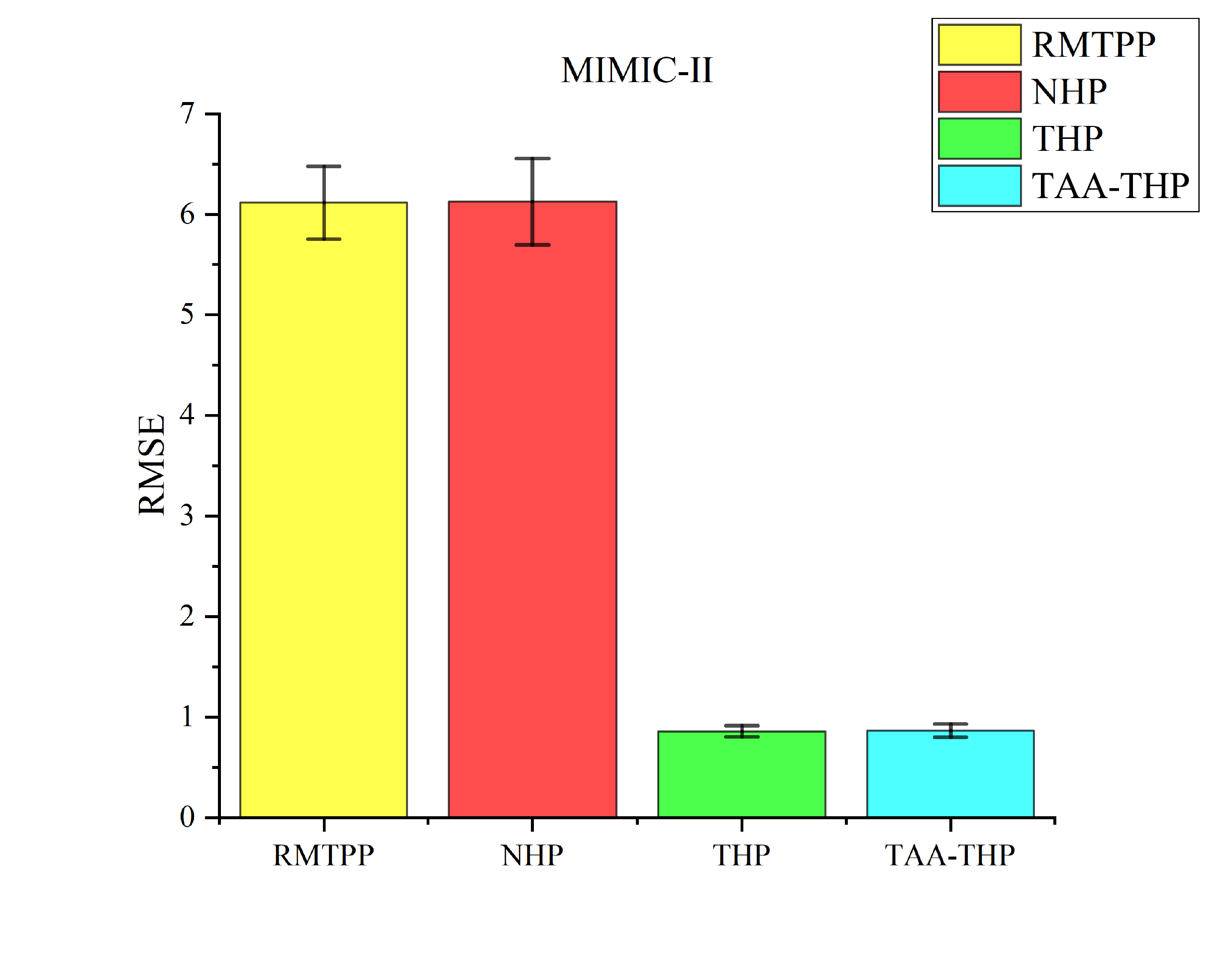}
	}
	\quad
	\subfigure[]{
		\includegraphics[width=4.5cm]{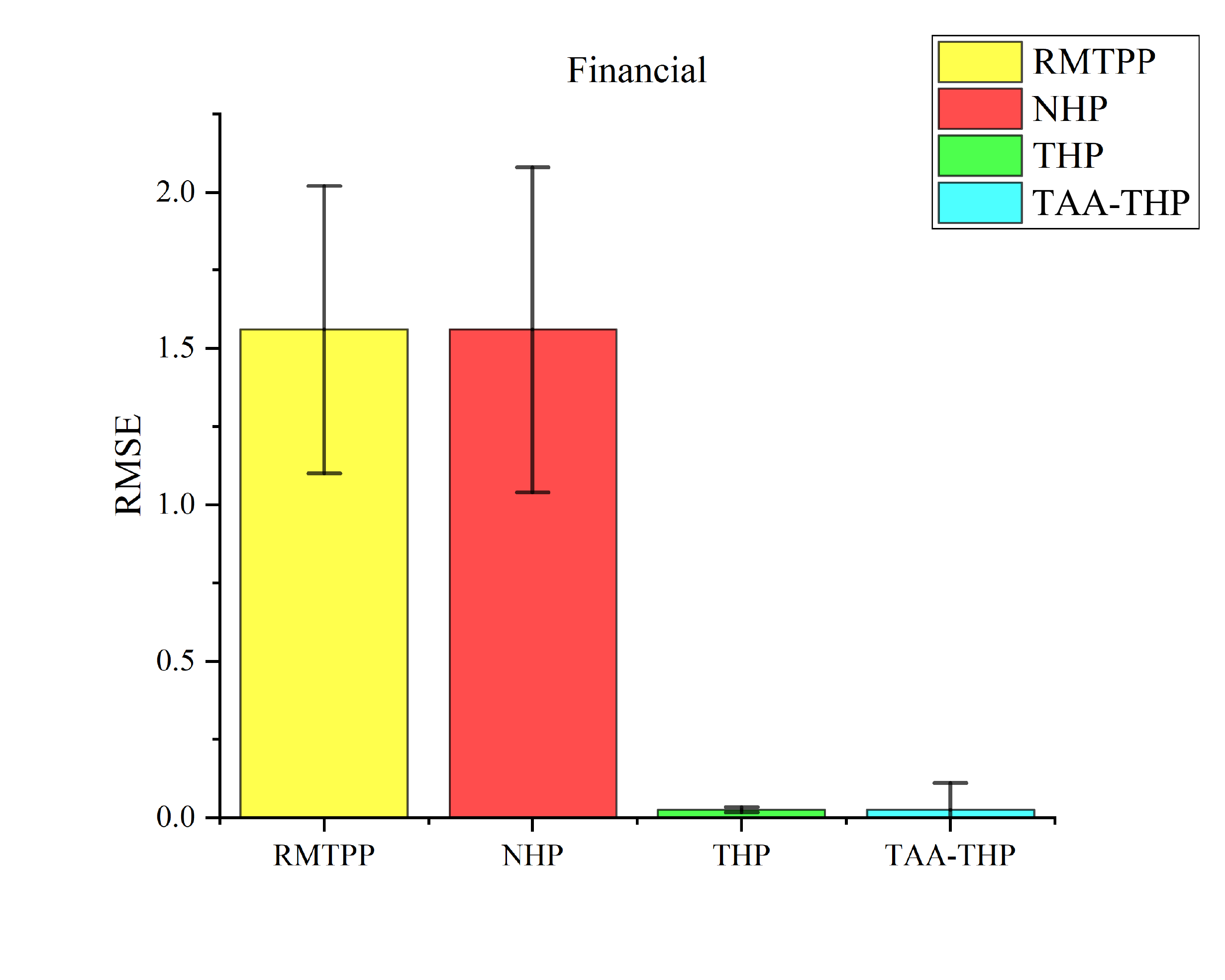}
	}
	\subfigure[]{
	\includegraphics[width=4.5cm]{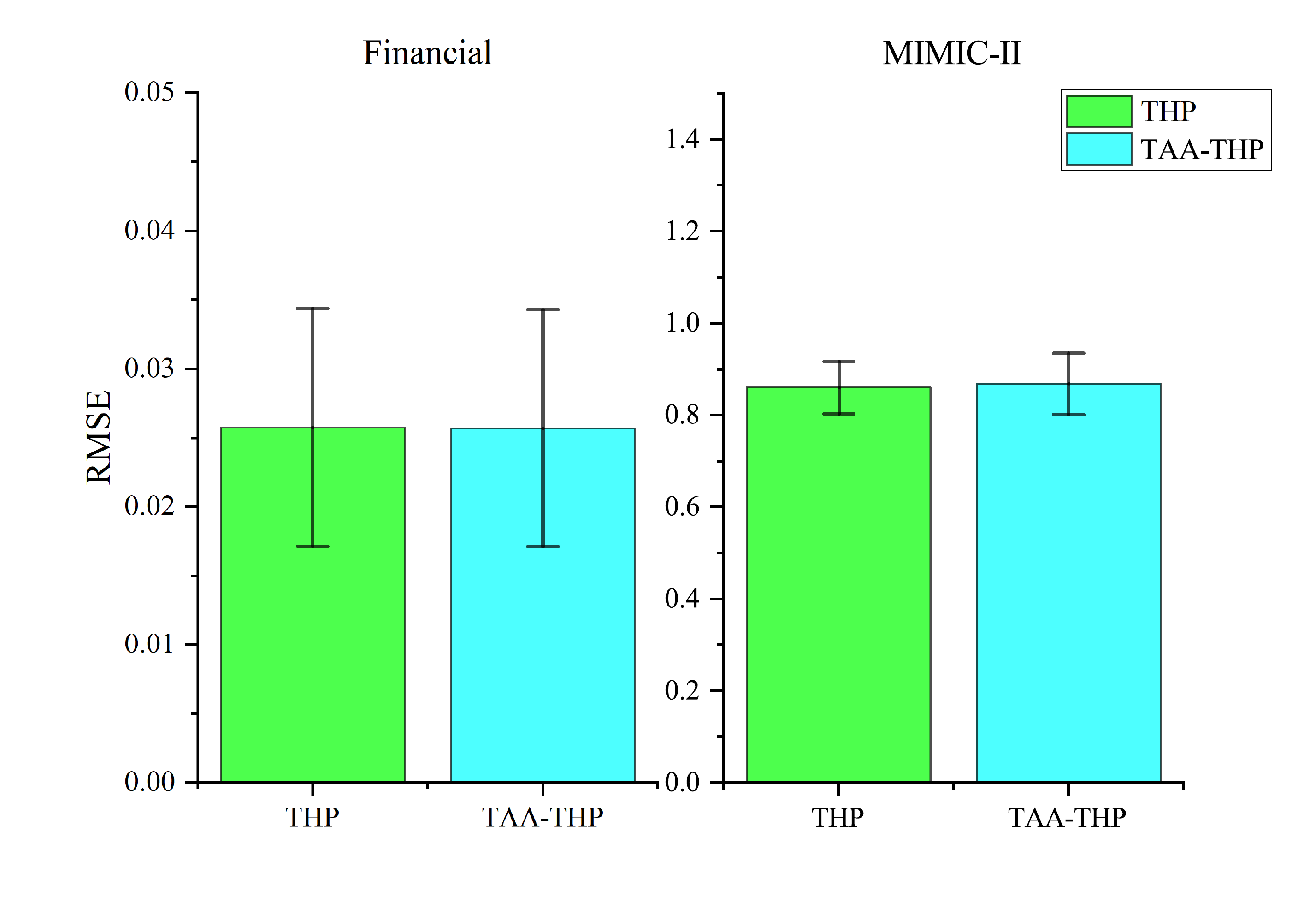}
	}
	\caption{RMSE for existing baselines and TAA-THP. Based on the five times train-dev-test partition, five experiments are performed on each dataset, and then the mean and standard deviation of different models are obtained.}
\end{figure}

From Fig. 4 and Table 6, we can find out that the RMSE of TAA-THP is generally smaller than the ones of the existing baselines. The experimental results for Loglike, accuracy, and RMSE has the obvious improvement than the existing baselines on the different scenes, for example, for the MIMIC-II dataset the number of the event types is 75, and the average length of sequences is only 4, for the Retweets dataset the number of the event types is 3, and the average length of sequences is 264, while for StackOverflow dataset the number of the event types is 22, and the average length of sequences is 736, both of them are relatively big, and for Financial dataset, the number of event types is only 2, and the average length of sequences is 2074. The characteristics of these datasets are obvious differences, our model shows a certain degree of robustness, and we can conclude that the TAA-THP can effectively capture the short-term and long-term dependencies between events.

\subsection{Ablation study}

So as to figure out the specific influence of the introduction of temporal augmented attention, we conduct the ablation study on StackOverflow, MIMIC-II, and Financial datasets. If we remove the temporal augmented attention in TAA-THP network architecture, the only difference between it and traditional dot-product attention is the bias ${\bm{b}}_{lq} $ . Therefore, here, we temporarily name this attention structure as biased attention, and the dot-product operation is shown as Eq.\ref{eq18}:

\begin{equation}
	\label{eq18}
{\bf{A}}_l  = Softmax\left[ {mask\left( {\frac{{\left( {{\bf{Q}}_l  + {\bf{b}}_{lq} } \right){\bf{K}}_l^T }}{{\sqrt {D_K } }}} \right)} \right]{\bf{V}}_l 
\end{equation}	

Compare with Eq.3, the term $\left( {{\bf{Q}}_l  + {\bf{b}}_{lt} } \right)\left( {{\bf{X}}^T {\bf{W}}_{Tem}^l } \right)^T $ is removed, which is the temporal attention, we compare the three evaluation criteria, e.g., Accuracy, RMSE, and Loglike on three datasets, the results are shown in Table 7. And to more intuitively compare the impact of temporal attention, we visualize the performance of models in Fig. 5, Fig. 6 and Fig. 7.

\begin{table}[!htbp]
	\centering
	\caption{The ablation experimental results of temporal attention.}
	\begin{tabular}{cccc}
		\hline
		Dataset                        & Evaluation criteria & Biased attention & TAA-THP \\ \hline
		\multirow{3}{*}{StackOverflow} & Accuracy            & 46.80            & 46.83   \\
		& RMSE                & 4.04             & 3.91    \\
		& Loglike             & -0.547           & -0.545  \\ \hline
		\multirow{3}{*}{MIMIC-II}      & Accuracy            & 84.00            & 84.41   \\
		& RMSE                & 0.872            & 0.868   \\
		& Loglike             & -0.129           & -0.111  \\ \hline
		\multirow{3}{*}{Financial}     & Accuracy            & 62.38            & 62.44   \\
		& RMSE                & 0.02573          & 0.02570 \\
		& Loglike             & -1.03            & -1.08   \\ \hline
	\end{tabular}
\end{table}

\begin{figure}[!htbp]
	
	\centering
	\subfigure[]{
		\includegraphics[width=4.5cm]{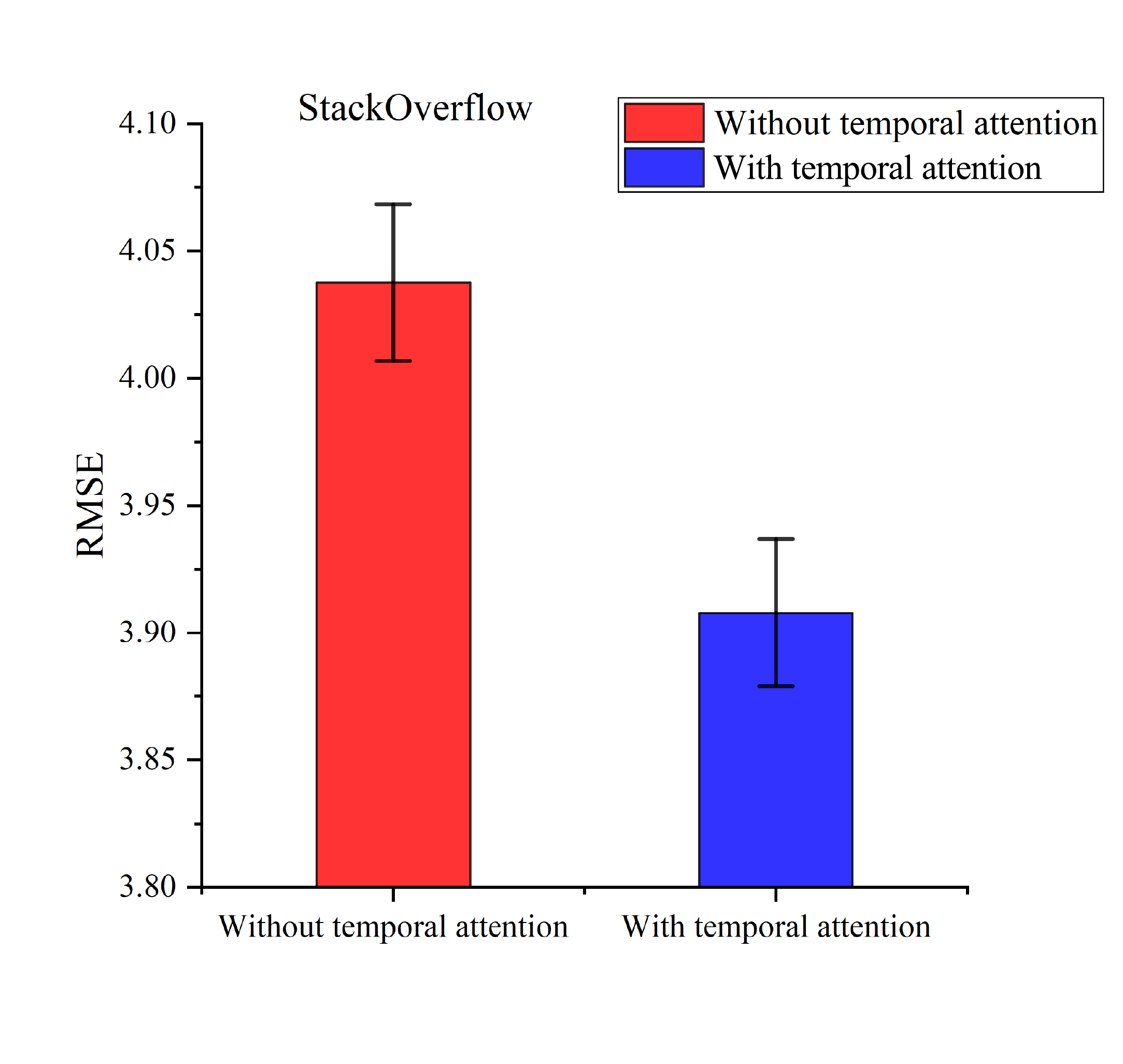}
	}
	\quad
	\subfigure[]{
		\includegraphics[width=4.5cm]{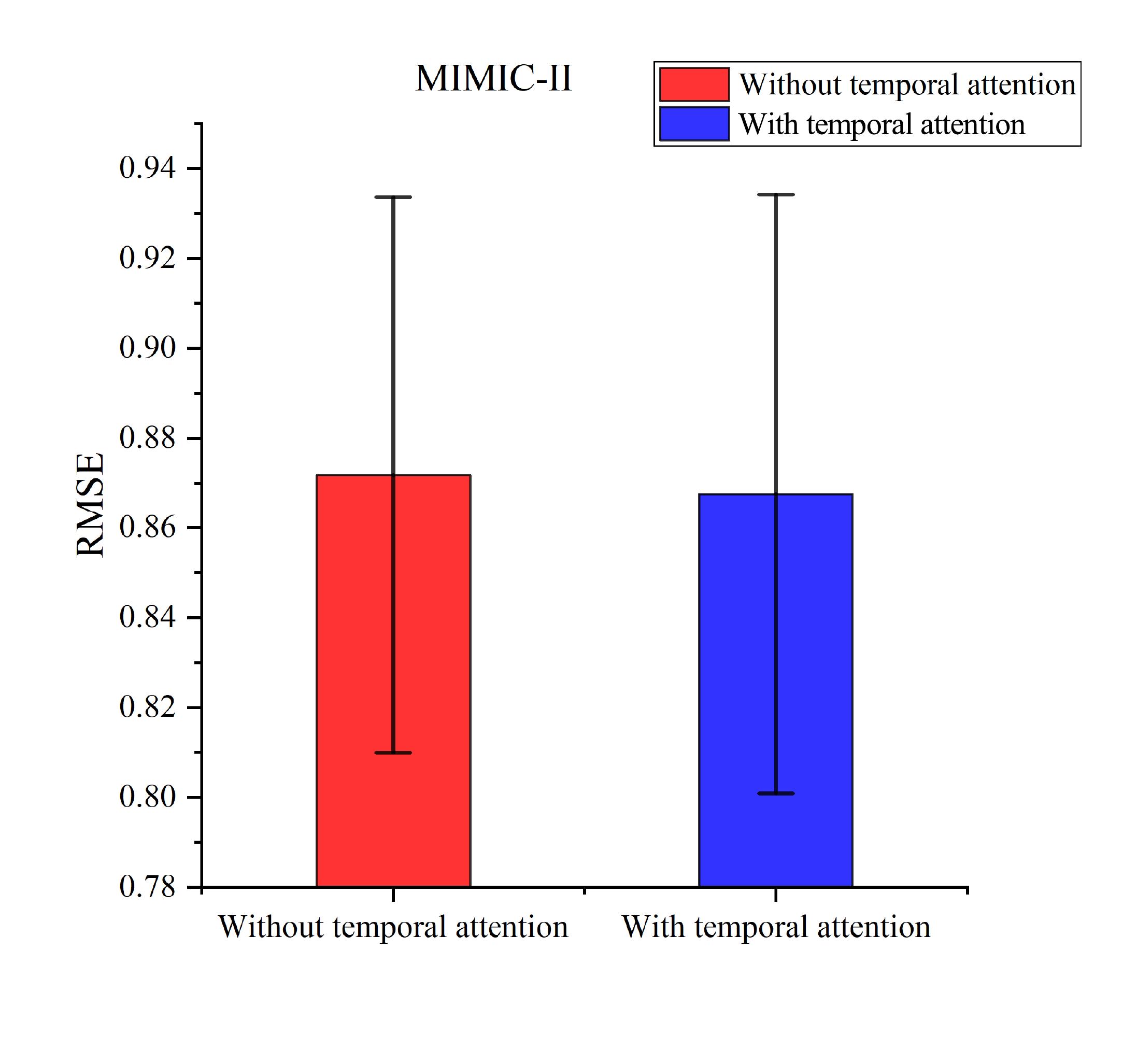}
	}
	\quad
	\subfigure[]{
		\includegraphics[width=4.5cm]{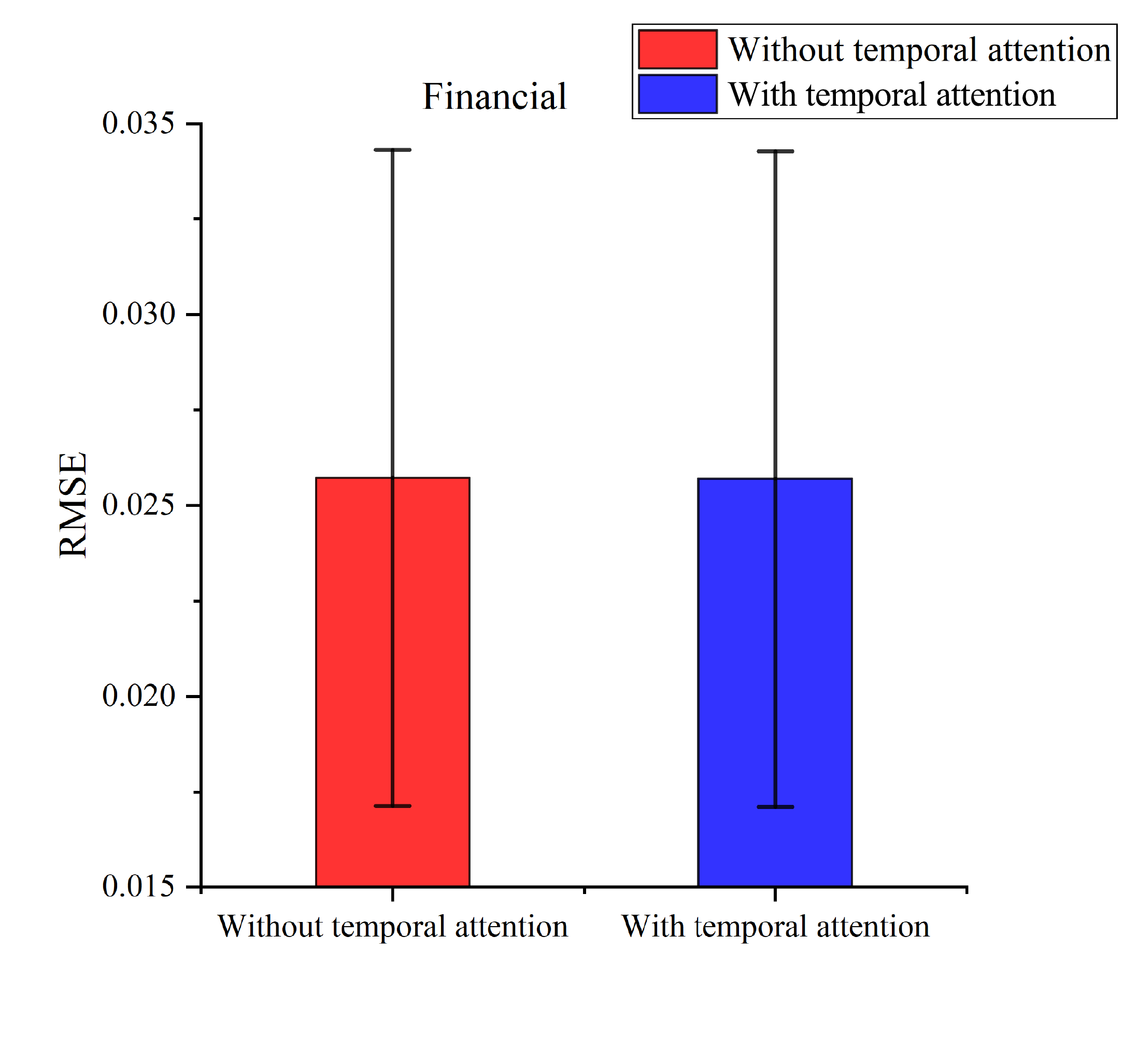}
	}
	\caption{in terms of Loglike, visualization of ablation experimental results with and without temporal attention.}
\end{figure}

\begin{figure}[!htbp]
	
	\centering
	\subfigure[]{
		\includegraphics[width=4.5cm]{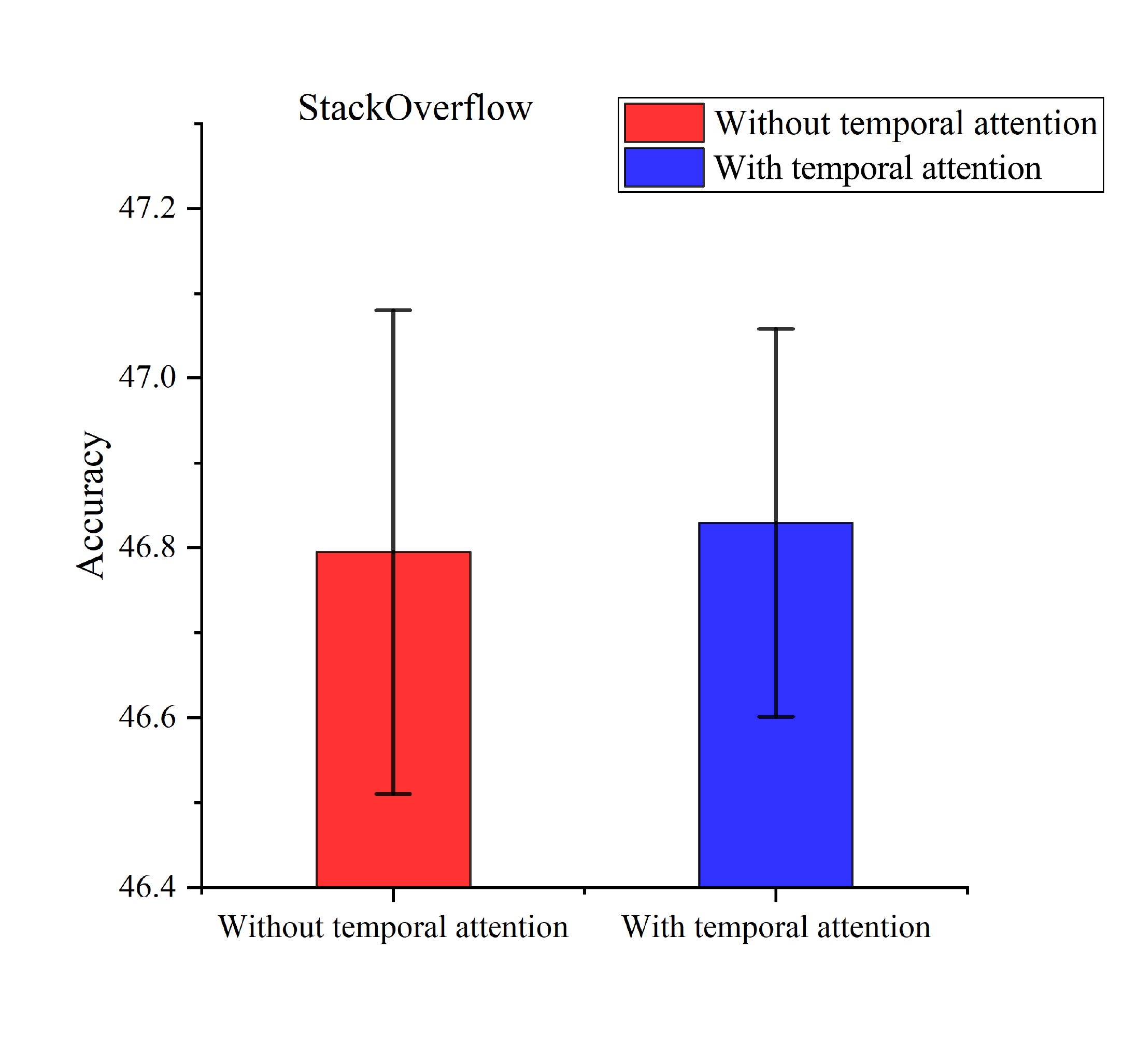}
	}
	\quad
	\subfigure[]{
		\includegraphics[width=4.5cm]{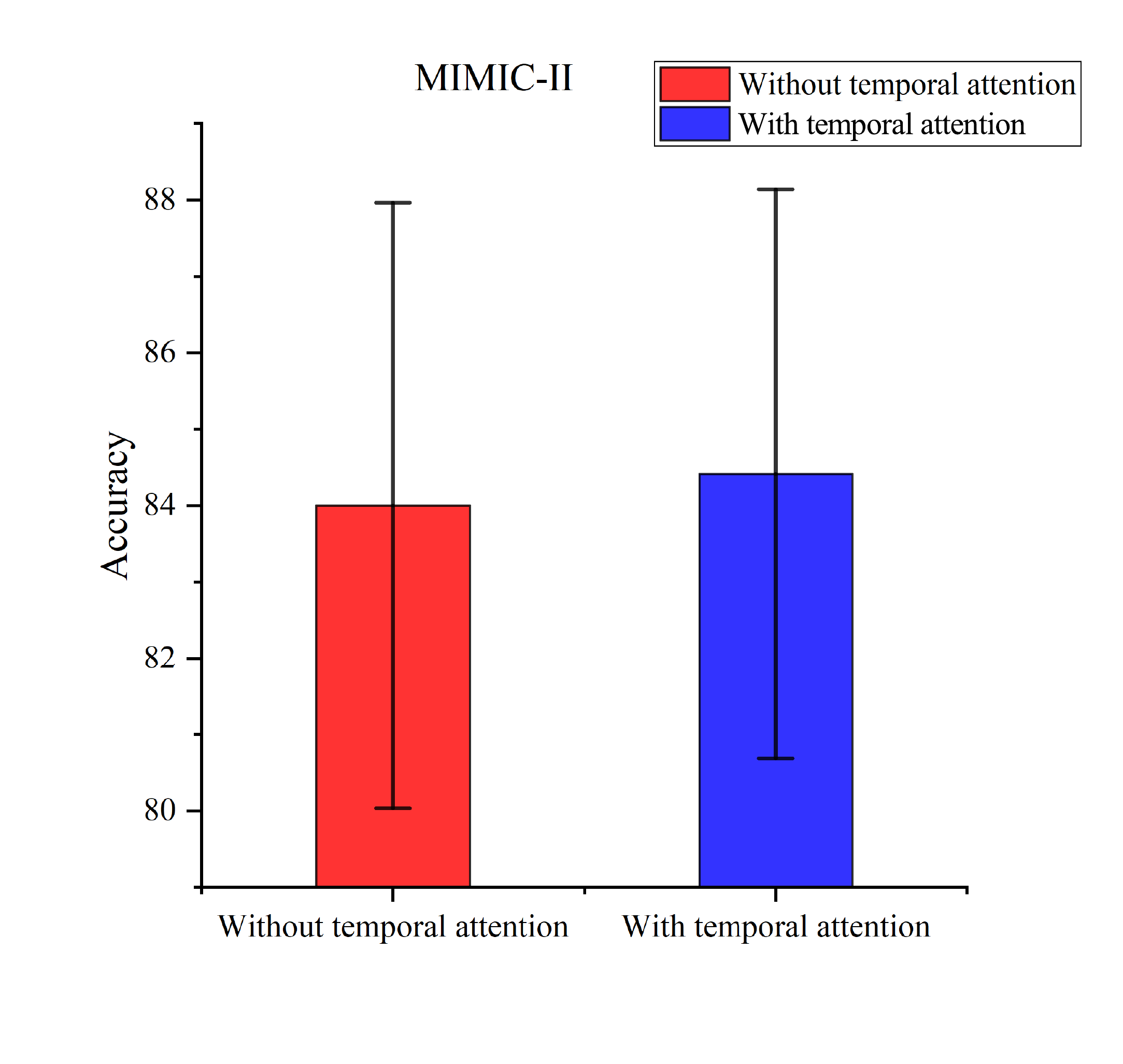}
	}
	\quad
	\subfigure[]{
		\includegraphics[width=4.5cm]{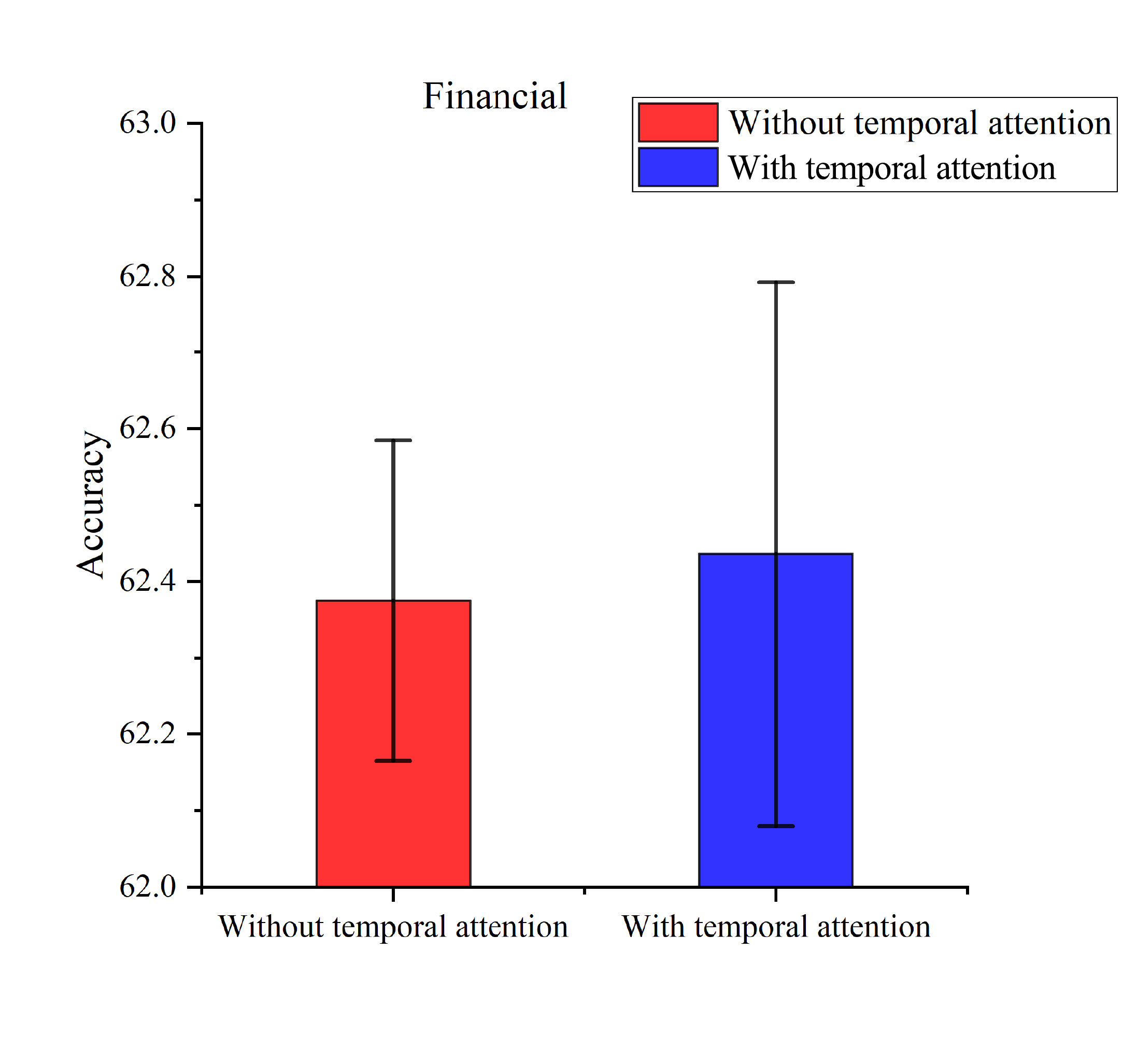}
	}
	\caption{in terms of RMSE criteria, visualization of ablation experimental results with and without temporal attention. }
\end{figure}

\begin{figure}[!htbp]

	\centering
	\subfigure[]{
		\includegraphics[width=4.5cm]{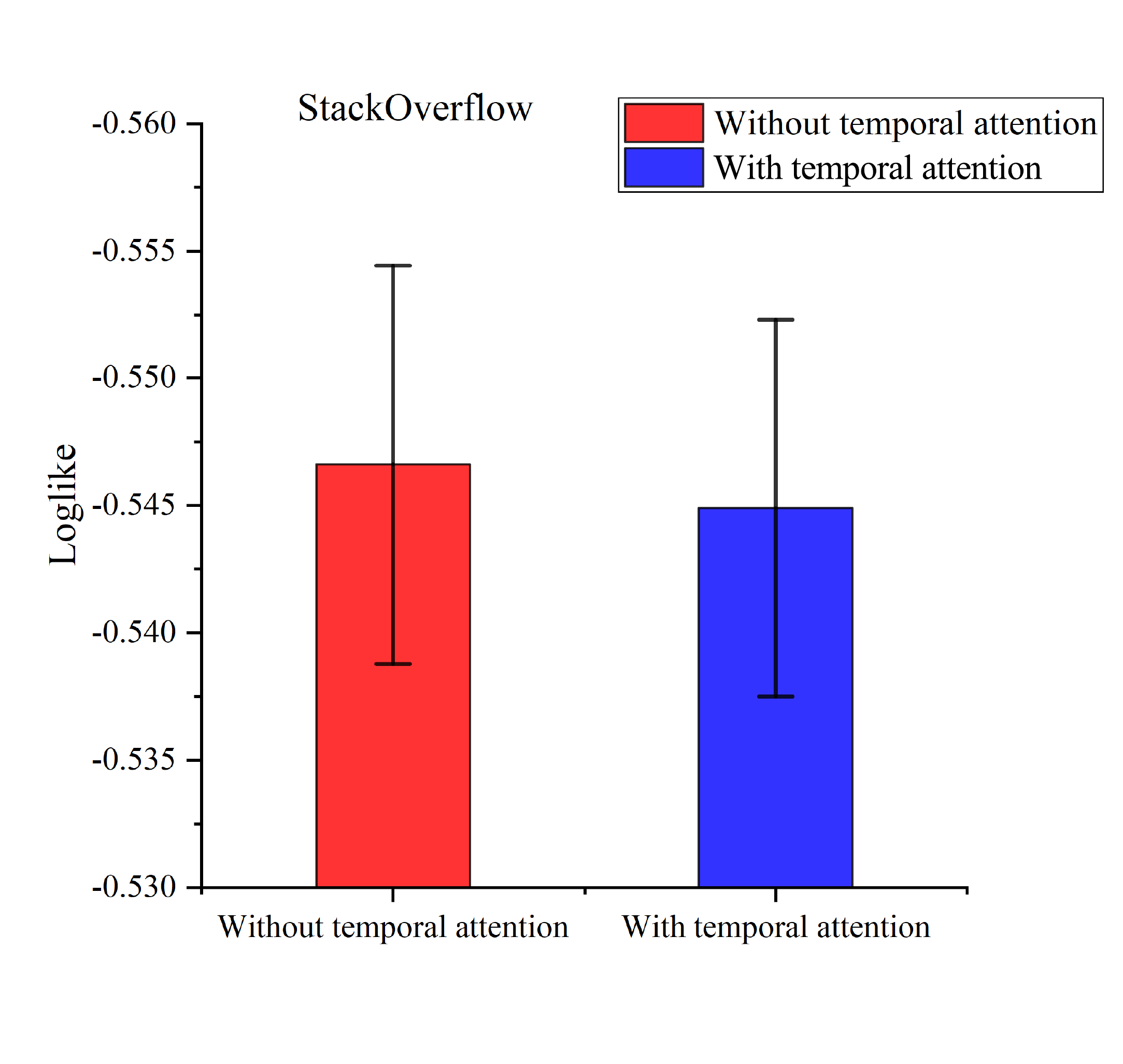}
	}
	\quad
	\subfigure[]{
		\includegraphics[width=4.5cm]{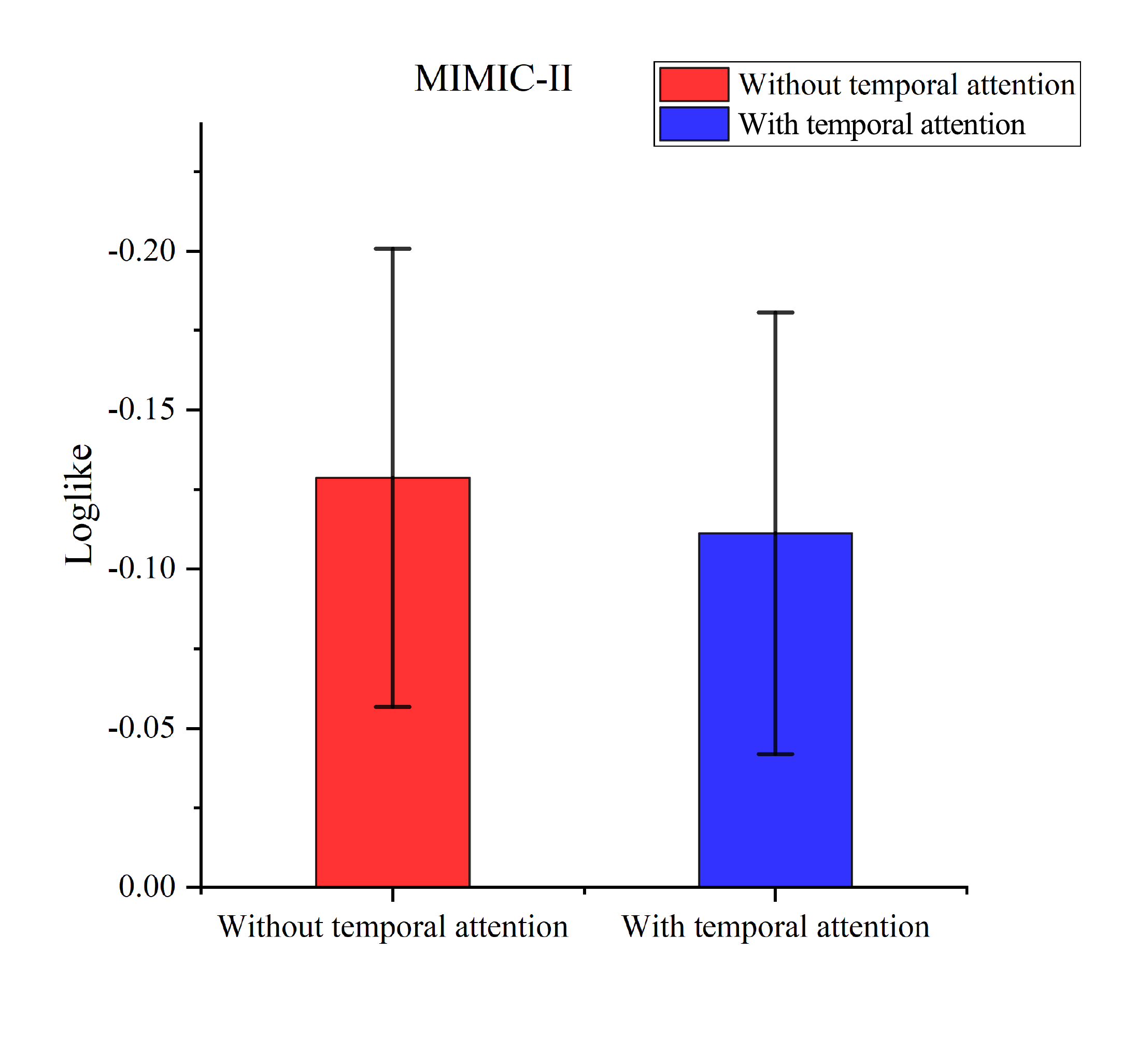}
	}
	\quad
	\subfigure[]{
		\includegraphics[width=4.5cm]{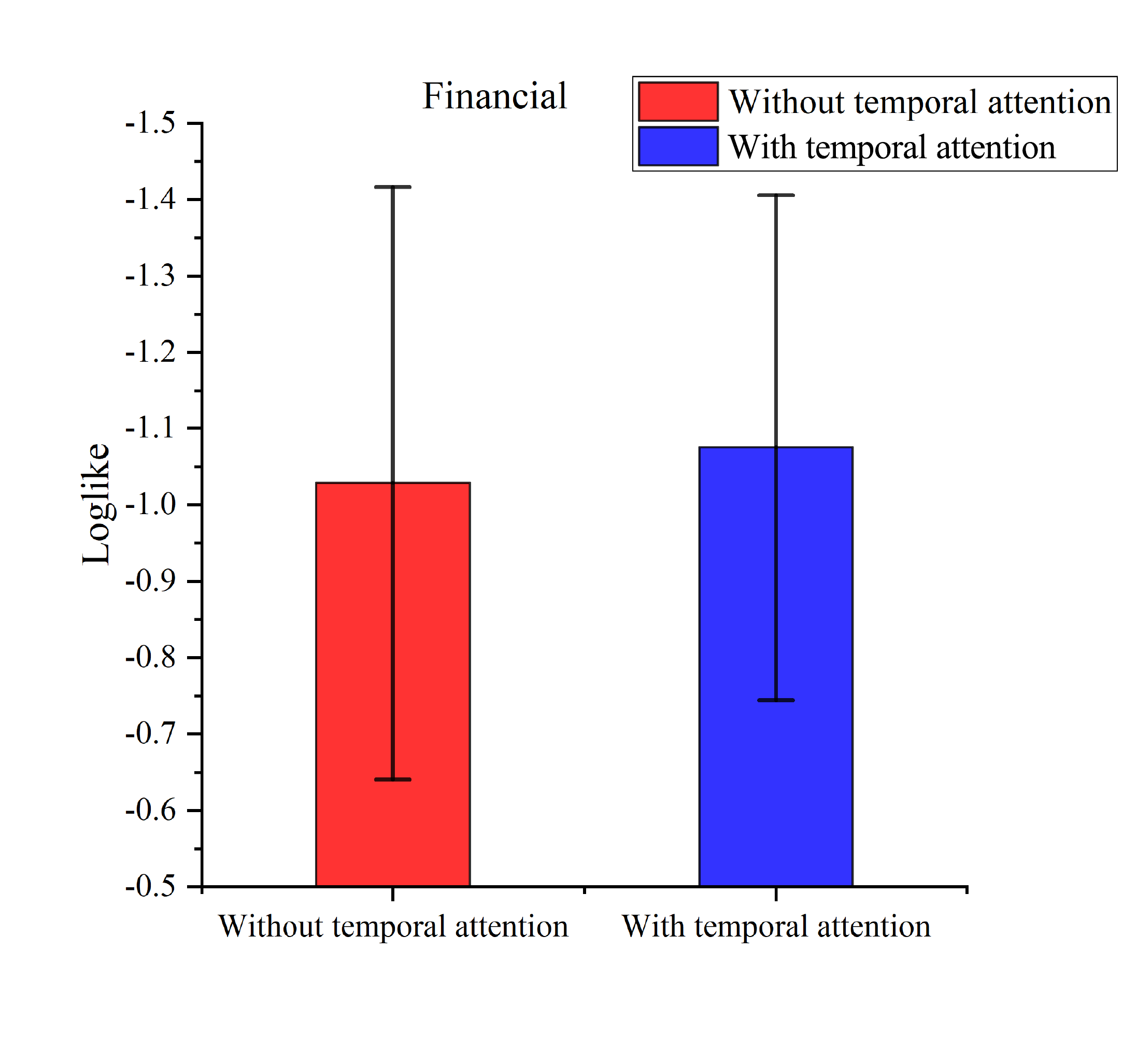}
	}
	\caption{in terms of prediction accuracies, visualization of ablation experimental results with and without temporal attention}
\end{figure}

From Fig. 5 and Table 7, we can see that with removing of temporal attention, the Loglikes on the datasets become smaller overall, which verifies the effectiveness of introducing temporal attention mechanism. Meanwhile, we can find out that the Loglike and accuracies of Biased attention are still higher than the existing baselines, we guess that this is the effect of bias ${\bm{b}}_{lq}$, which can keep most of the neurons in the model activated. 

Fig. 6, Fig. 7, and Table 7 show the prediction performance of models, we find that the lack of temporal attention reduces the prediction performance of event occurring time and type of event. This result confirms the merit of introducing temporal attention. Moreover, after removing temporal attention from the TAA-THP model, the prediction accuracies of the event will decrease, which can test and verify the hypothesis we put forward in subsection 5.3, the additional temporal attention indeed helps to decouple the simple concatenating of event encoding and time position encoding. In our standpoint, when we introduce temporal attention, it passes through a linear transformation ${\bm{X}}^T {\bm{W}}_{Tem}^l $ which is depicted in Eq.\ref{eq3}, which is similar to the training mechanism of the residual network. During the model training, the model can spontaneously decide whether the temporal information will feed into the multi-head attention, how much information is actually sent over the multi-head attention, it depends on the value of elements of $
{\bm{W}}_{Tem}^l $
 , through the experimental results, we see that the model with temporal attention performs better, which confirms that temporal attention does exist in the model in its own right, and it plays an effective and positive role.

\section{Conclusions and future works}

In this paper, we come up with a new structure of the transformer Hawkes process. We improve the traditional dot-product attention mechanism. We propose to introduce temporal attention to the encoder of the transformer. The experimental results confirm that the additional introduction of temporal attention significantly improves the performance, especially in the predicting error for the occurring time and type of next event, and the overall likelihood function value on the sequence of events. In addition, through additional ablation study, we verify that the introduction of temporal attention is indeed effective. And in the future, we hope to propose a more valuable model based on the transformer and incorporate another generative neural network model to acquire better performance.

\bibliography{mybibfile}

\begin{thebibliography}{10}
\expandafter\ifx\csname url\endcsname\relax
  \def\url#1{\texttt{#1}}\fi
\expandafter\ifx\csname urlprefix\endcsname\relax\def\urlprefix{URL }\fi
\expandafter\ifx\csname href\endcsname\relax
  \def\href#1#2{#2} \def\path#1{#1}\fi

\bibitem{1ogata1981lewis}
Y.~Ogata, On lewis' simulation method for point processes, IEEE Transactions on
  Information Theory 27~(1) (1981) 23--31.

\bibitem{2johnson2016mimic}
A.~E. Johnson, T.~J. Pollard, L.~Shen, H.~L. Li-Wei, M.~Feng, M.~Ghassemi,
  B.~Moody, P.~Szolovits, L.~A. Celi, R.~G. Mark, Mimic-iii, a freely
  accessible critical care database, Scientific data 3~(1) (2016) 1--9.

\bibitem{3mohler2018improving}
G.~Mohler, J.~Carter, R.~Raje, Improving social harm indices with a modulated
  hawkes process, International Journal of Forecasting 34~(3) (2018) 431--439.

\bibitem{4zhang2020survival}
L.-n. Zhang, J.-w. Liu, X.~Zuo, Survival analysis of failures based on hawkes
  process with weibull base intensity, Engineering Applications of Artificial
  Intelligence 93 (2020) 103709.

\bibitem{5luo2014you}
D.~Luo, H.~Xu, H.~Zha, J.~Du, R.~Xie, X.~Yang, W.~Zhang, You are what you watch
  and when you watch: Inferring household structures from iptv viewing data,
  IEEE Transactions on Broadcasting 60~(1) (2014) 61--72.

\bibitem{6zhao2015seismic}
Q.~Zhao, M.~A. Erdogdu, H.~Y. He, A.~Rajaraman, J.~Leskovec, Seismic: A
  self-exciting point process model for predicting tweet popularity, in:
  Proceedings of the 21th ACM SIGKDD International Conference on Knowledge
  Discovery and Data Mining, ACM, 2015, pp. 1513--1522.

\bibitem{7daley2007introduction}
D.~J. Daley, D.~Vere-Jones, An introduction to the theory of point processes:
  volume II: general theory and structure, Springer Science \& Business Media,
  2007.

\bibitem{8hawkes1971spectra}
A.~G. Hawkes, Spectra of some self-exciting and mutually exciting point
  processes, Biometrika 58~(1) (1971) 83--90.

\bibitem{9reynaud2010adaptive}
P.~Reynaud-Bouret, S.~Schbath, et~al., Adaptive estimation for hawkes
  processes; application to genome analysis, The Annals of Statistics 38~(5)
  (2010) 2781--2822.

\bibitem{10kobayashi2016tideh}
R.~Kobayashi, R.~Lambiotte, Tideh: Time-dependent hawkes process for predicting
  retweet dynamics, arXiv preprint arXiv:1603.09449 (2016).

\bibitem{11xu2016learning}
H.~Xu, M.~Farajtabar, H.~Zha, Learning granger causality for hawkes processes,
  in: International Conference on Machine Learning, 2016, pp. 1717--1726.

\bibitem{12zhou2013learning}
K.~Zhou, H.~Zha, L.~Song, Learning social infectivity in sparse low-rank
  networks using multi-dimensional hawkes processes, in: Artificial
  Intelligence and Statistics, PMLR, 2013, pp. 641--649.

\bibitem{13du2016recurrent}
N.~Du, H.~Dai, R.~Trivedi, U.~Upadhyay, M.~Gomez-Rodriguez, L.~Song, Recurrent
  marked temporal point processes: Embedding event history to vector, in:
  Proceedings of the 22nd ACM SIGKDD International Conference on Knowledge
  Discovery and Data Mining, 2016, pp. 1555--1564.

\bibitem{14mei2017neural}
H.~Mei, J.~M. Eisner, The neural hawkes process: A neurally self-modulating
  multivariate point process, in: Advances in Neural Information Processing
  Systems, 2017, pp. 6754--6764.

\bibitem{15xiao2017modeling}
S.~Xiao, J.~Yan, X.~Yang, H.~Zha, S.~M. Chu, Modeling the intensity function of
  point process via recurrent neural networks, in: Thirty-First AAAI Conference
  on Artificial Intelligence, 2017.

\bibitem{16bengio1994learning}
Y.~Bengio, P.~Simard, P.~Frasconi, Learning long-term dependencies with
  gradient descent is difficult, IEEE transactions on neural networks 5~(2)
  (1994) 157--166.

\bibitem{17pascanu2013difficulty}
R.~Pascanu, T.~Mikolov, Y.~Bengio, On the difficulty of training recurrent
  neural networks, in: International conference on machine learning, 2013, pp.
  1310--1318.

\bibitem{18DBLP:journals/corr/BahdanauCB14}
D.~Bahdanau, K.~Cho, Y.~Bengio, \href{http://arxiv.org/abs/1409.0473}{Neural
  machine translation by jointly learning to align and translate}, in:
  Y.~Bengio, Y.~LeCun (Eds.), 3rd International Conference on Learning
  Representations, {ICLR} 2015, San Diego, CA, USA, May 7-9, 2015, Conference
  Track Proceedings, 2015.
\newline\urlprefix\url{http://arxiv.org/abs/1409.0473}

\bibitem{19vaswani2017attention}
A.~Vaswani, N.~Shazeer, N.~Parmar, J.~Uszkoreit, L.~Jones, A.~N. Gomez,
  {\L}.~Kaiser, I.~Polosukhin, Attention is all you need, in: Advances in
  neural information processing systems, 2017, pp. 5998--6008.

\bibitem{20yu2016automatic}
D.~Yu, L.~Deng, Automatic Speech Recognition., Springer, 2016.

\bibitem{21koehn2009statistical}
P.~Koehn, Statistical machine translation, Cambridge University Press, 2009.

\bibitem{22girdhar2019video}
R.~Girdhar, J.~Carreira, C.~Doersch, A.~Zisserman, Video action transformer
  network, in: Proceedings of the IEEE/CVF Conference on Computer Vision and
  Pattern Recognition, 2019, pp. 244--253.

\bibitem{23zhang2020self}
Q.~Zhang, A.~Lipani, O.~Kirnap, E.~Yilmaz, Self-attentive hawkes process, in:
  International Conference on Machine Learning, PMLR, 2020, pp. 11183--11193.

\bibitem{24DBLP:conf/icml/ZuoJLZZ20}
S.~Zuo, H.~Jiang, Z.~Li, T.~Zhao, H.~Zha,
  \href{http://proceedings.mlr.press/v119/zuo20a.html}{Transformer hawkes
  process}, in: Proceedings of the 37th International Conference on Machine
  Learning, {ICML} 2020, 13-18 July 2020, Virtual Event, Vol. 119 of
  Proceedings of Machine Learning Research, {PMLR}, 2020, pp. 11692--11702.
\newline\urlprefix\url{http://proceedings.mlr.press/v119/zuo20a.html}

\bibitem{25dai2019transformer}
Z.~Dai, Z.~Yang, Y.~Yang, J.~Carbonell, Q.~V. Le, R.~Salakhutdinov,
  Transformer-xl: Attentive language models beyond a fixed-length context,
  arXiv preprint arXiv:1901.02860 (2019).

\bibitem{26al2019character}
R.~Al-Rfou, D.~Choe, N.~Constant, M.~Guo, L.~Jones, Character-level language
  modeling with deeper self-attention, in: Proceedings of the AAAI Conference
  on Artificial Intelligence, Vol.~33, 2019, pp. 3159--3166.

\bibitem{27yang2017online}
Y.~Yang, J.~Etesami, N.~He, N.~Kiyavash, Online learning for multivariate
  hawkes processes, Advances in Neural Information Processing Systems 30 (2017)
  4937--4946.

\bibitem{28hawkes2018hawkes}
A.~G. Hawkes, Hawkes processes and their applications to finance: a review,
  Quantitative Finance 18~(2) (2018) 193--198.

\bibitem{29hansen2015lasso}
N.~R. Hansen, P.~Reynaud-Bouret, V.~Rivoirard, et~al., Lasso and probabilistic
  inequalities for multivariate point processes, Bernoulli 21~(1) (2015)
  83--143.

\bibitem{30DBLP:conf/iclr/DehghaniGVUK19}
M.~Dehghani, S.~Gouws, O.~Vinyals, J.~Uszkoreit, L.~Kaiser,
  \href{https://openreview.net/forum?id=HyzdRiR9Y7}{Universal transformers},
  in: 7th International Conference on Learning Representations, {ICLR} 2019,
  New Orleans, LA, USA, May 6-9, 2019, OpenReview.net, 2019.
\newline\urlprefix\url{https://openreview.net/forum?id=HyzdRiR9Y7}

\bibitem{31graves2016adaptive}
A.~Graves, Adaptive computation time for recurrent neural networks, arXiv
  preprint arXiv:1603.08983 (2016).

\bibitem{32wang2019language}
C.~Wang, M.~Li, A.~J. Smola, Language models with transformers, arXiv preprint
  arXiv:1904.09408 (2019).

\bibitem{33robert2013monte}
C.~Robert, G.~Casella, Monte Carlo statistical methods, Springer Science \&
  Business Media, 2013.

\bibitem{34stoer2013introduction}
J.~Stoer, R.~Bulirsch, Introduction to numerical analysis, Vol.~12, Springer
  Science \& Business Media, 2013.

\bibitem{35kingma2014adam}
D.~P. Kingma, J.~Ba, Adam: A method for stochastic optimization, arXiv preprint
  arXiv:1412.6980 (2014).

\bibitem{36leskovec2014snap}
J.~Leskovec, A.~Krevl, Snap datasets: Stanford large network dataset collection
  (2014).

\end{thebibliography}

\end{document}